\documentclass[letterpaper]{article} 
\usepackage{aaai24}  
\usepackage{times}  
\usepackage{helvet}  
\usepackage{courier}  
\usepackage[hyphens]{url}  
\usepackage{graphicx} 
\usepackage{natbib}  
\usepackage{caption} 
\usepackage{algorithm}

\usepackage{newfloat}
\usepackage{listings}
\DeclareCaptionStyle{ruled}{labelfont=normalfont,labelsep=colon,strut=off} 
\floatstyle{ruled}
\newfloat{listing}{tb}{lst}{}
\floatname{listing}{Listing}
\nocopyright

\usepackage{amssymb}
\usepackage{adjustbox}     
\usepackage{microtype}
\usepackage{graphicx}
\usepackage{amsmath}
\usepackage{lipsum}
\usepackage{algpseudocode}
\usepackage{amsfonts}
\usepackage{subcaption}
\usepackage{comment}
\usepackage{url}
\usepackage{multirow}
\usepackage{multirow}
\usepackage{booktabs}
\DeclareUnicodeCharacter{2212}{-}
\usepackage{makecell}

\usepackage{float}

\usepackage{xcolor}

\newcommand{\answerTODO}[1]{\textcolor{red}{#1}} 

\title{Discovering Latent Themes in Social Media Messaging: A Machine-in-the-Loop Approach Integrating LLMs}

\author {
    Tunazzina Islam,
    Dan Goldwasser
}
\affiliations {
    Department of Computer Science, Purdue University, West Lafayette, IN 47907, USA.
    \\
    islam32@purdue.edu, dgoldwas@purdue.edu
}

\usepackage{bibentry}

\begin{document}

\maketitle

\begin{abstract}
Grasping the themes of social media content is key to understanding the narratives that influence public opinion and behavior. The thematic analysis goes beyond traditional topic-level analysis, which often captures only the broadest patterns, providing deeper insights into specific and actionable themes such as ``public sentiment towards vaccination", ``political discourse surrounding climate policies," etc.
In this paper, we introduce a novel approach to uncovering latent themes in social media messaging. Recognizing the limitations of the traditional topic-level analysis, which tends to capture only overarching patterns, this study emphasizes the need for a finer-grained, theme-focused exploration. Traditional theme discovery methods typically involve manual processes and a human-in-the-loop approach. While valuable, these methods face challenges in scalability, consistency, and resource intensity in terms of time and cost. To address these challenges, we propose a \textbf{machine-in-the-loop} approach that leverages the advanced capabilities of Large Language Models (LLMs). 
This approach facilitates a deeper investigation into the social media discourse, revealing a variety of themes with distinct characteristics and relevance. It provides a detailed understanding of underlying nuances and efficiently maps texts to these themes, enhancing our insight into social media messaging.

To demonstrate our approach, we apply our framework to contentious topics, such as \textit{climate debate} and \textit{vaccine debate}. We use two publicly available datasets: (1) the climate campaigns dataset of $21k$ Facebook ads and (2) the COVID-19 vaccine campaigns dataset of $9k$ Facebook ads. Our quantitative and qualitative analysis shows that our methodology yields more accurate and interpretable results compared to the baselines. Our results not only demonstrate the effectiveness of our approach in uncovering latent themes but also illuminate how these themes are tailored for demographic targeting in social media contexts. Additionally, our work sheds light on the dynamic nature of social media, revealing the shifts in the thematic focus of messaging in response to real-world events.

\end{abstract}

\section{Introduction}
The social media landscape is highly diverse and dynamic, serving as a battleground for various interest groups, including politicians, advertisers, and other stakeholders. These entities strategically use these platforms for \textit{microtargeting} \cite{hersh2015,barbu2014advertising}, a process that utilizes data-driven techniques to target potential users and advance their interests. Microtargeting capitalizes on the rich user data collected by social networks, making the understanding of this targeted messaging a significant challenge.

In recent years, social media has emerged as a fertile ground for diverse discussions and expressions, encompassing a broad spectrum of topics and themes.
In this context, a `topic' in social media content represents a broad category that includes various specific and nuanced `themes'. The topic acts as an umbrella term under which different themes fall. For example, under the `Climate Change' topic, a theme might be `Renewable Energy Solutions'.  Identifying themes within a topic is to explore this specific area in depth, thereby uncovering the detailed and granular nuances that define the broader category. Given the dynamic nature of social media, accurately discerning and understanding the diverse themes within a broader topic is a complex and continuously evolving task.


Traditionally, thematic analysis (TA) has been the method of choice for this task, serving as a qualitative research technique focused on pattern identification, where themes emerging from the data guide further analysis \cite{roberts2019attempting,vaismoradi2016theme,vaismoradi2013content,braun2012thematic,braun2006using,tuckett2005applying}. However, with the explosion of textual data in the digital age, there's a growing inclination towards computational methods for text analysis \cite{sivarajah2017critical,jagadish2014big,fan2014challenges}.

Among these, supervised text analysis methods like Support Vector Machines (SVMs) \cite{joachims1998text}, Naive Bayes classifiers \cite{mccallum1998comparison}, and Neural Networks \cite{lecun2015deep} have been effective in theme categorization. These methods use pre-labeled datasets to learn and identify patterns, aiding in the categorization of new data. However, they present challenges, notably the need for extensive labeled datasets, which can be resource-intensive to create. 
Moreover, their performance heavily relies on the quality and representativeness of the training data. If the training data is biased or lacks diversity, the model's ability to generalize and accurately identify themes in varied or new contexts is significantly compromised. 

On the other hand, unsupervised text analysis methods, such as Latent Dirichlet Allocation (LDA) \cite{blei2003latent} and non-negative matrix factorization (NMF) \cite{lee1999learning}, have gained popularity in theme discovery. These methods independently identify themes from data without the need for pre-labeled datasets. While useful, they also have limitations.
Given that topics in topic models are defined as distributions over words, they can be challenging to interpret or may lack coherence, complicating the extraction of meaningful themes \cite{roder2015exploring,lau2014machine,mimno2011optimizing,chang2009reading}. 
Recent advancements have seen the integration of human judgment through a human-in-the-loop approach, which, while enhancing accuracy, suffers from issues of scalability and high resource demand \cite{pacheco2022holistic,pacheco2022interactively,lund2017tandem,hu2014interactive}.
Recognizing this, our work explores a new approach, integrating the strengths of LLMs \cite{brown2020language} with a streamlined algorithmic process to efficiently discover and characterize themes in social media messaging.
This exploration leads us to the following research questions (RQ), which are crucial for assessing the potential of LLMs to enhance our analysis by providing deeper and more nuanced insights:
\begin{enumerate}
    \item \textbf{RQ1:} Can LLMs determine if two given texts, without prior knowledge of existing codes, are discussing the same topic? 
    
    \item \textbf{RQ2:} If provided with a definition of a theme, can LLMs successfully categorize other texts under that specific theme? 

\end{enumerate}

    

\begin{figure}[htbp]
  \centering  
  \includegraphics[width= 1\columnwidth]{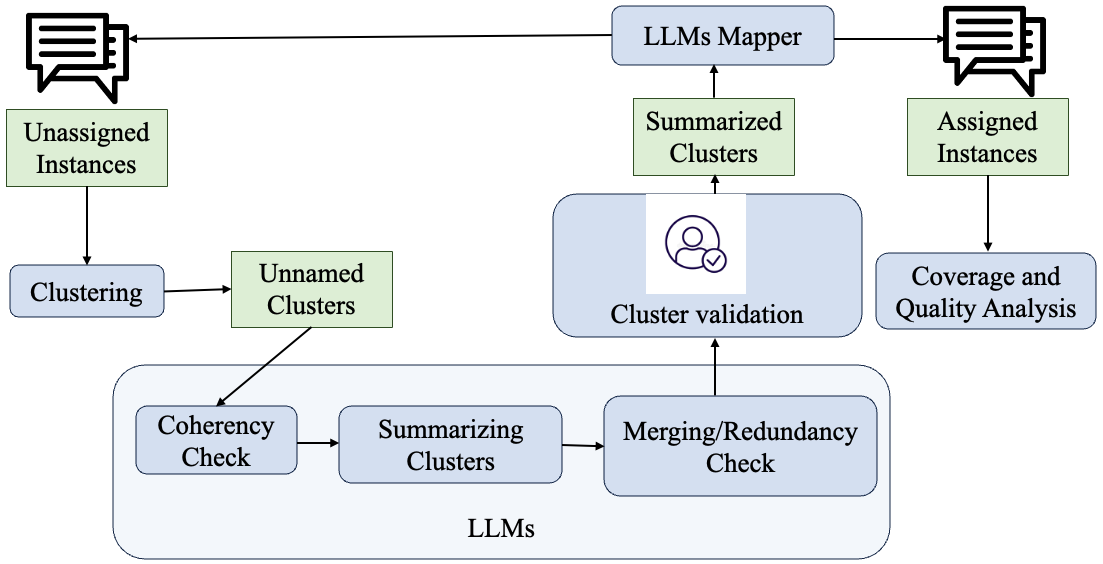}
  \caption{Framework overview.}
    \label{fig:llm}
\end{figure}

To enhance theme discovery, we propose
an iterative three-step \textit{machine-in-the-loop} framework. We define large textual collections as repositories of textual
instances (e.g., ads, tweets, posts, documents) where
each instance is potentially associated with a set of
annotated themes (seed sets from previous studies by \citet{islam2023analysis,islam2022understanding}). The seed sets of themes used in this paper are shown in Table \ref{tab:pre_thm}. 
A high-level overview of our method is illustrated in Figure \ref{fig:llm}.

\noindent \textbf{Firstly,} we initiate with candidate generation, a phase that includes clustering of instances using a clustering algorithm, ensuring consistency and coherence of the clusters using LLMs, and summarizing each cluster using LLMs. This phase also incorporates LLMs for redundancy checks and merging overlapping clusters, thus refining the cluster pool.

\noindent \textbf{Secondly,} we acknowledge the potential need for human validation. This optional phase allows for a critical assessment of the newly formed clusters, questioning their merging, description, and overall coherence. This step serves as a bridge, combining the computational efficiency of algorithms with the nuanced understanding of human judgment.

\noindent \textbf{Finally,} the assignment phase utilizes the power of LLMs in a few-shot prompting manner. This innovative approach determines if a new post/text aligns with the summarized clusters, thus assigning it to a relevant cluster. 

To showcase our approach, we examine two distinct case
studies: the \textbf{climate campaigns}
and the \textbf{COVID-19 vaccine campaigns} in the United States.
In each case, the qualitative researchers
apply different theories to ground theme discovery, each associated with a different set of concepts. For climate campaigns, the identification of themes is based on 
the energy industry and climate change-related stances \cite{islam2023analysis}. For the COVID-19 vaccine campaigns \cite{islam2022understanding}, the theme discovery is grounded using Moral Foundation Theory (MFT) \cite{haidt2007morality,haidt2004intuitive}. 
MFT abstracts over specific contentious topics to provide a general framework for understanding human morality. It suggests that there are at least six basic foundations—each with a positive and negative polarity—that explain the similarities and recurrent themes in morality across cultures.
These case studies were chosen for their significance to the computational social science (CSS) community, highlighting the challenges of targeted messaging in the digital era. From a machine learning perspective, they present unique challenges due to the specifics of the data, context, and themes involved.
Our contributions are summarized as follows:
\begin{enumerate}
\item  We suggest a \textbf{LLMs-in-the-loop} approach to uncover latent themes of social media messaging.
\item We conduct \textbf{quantitative} and \textbf{qualitative analysis} on real-world datasets to demonstrate the effectiveness of our proposed method. We show that our model outperforms the baselines.
\item Based on the uncovered themes, we
\textbf{analyze demographic targeting} on social media messaging. 
\item We demonstrate how the thematic emphasis of advertisements evolves in response to \textbf{real-world events}.
\end{enumerate}
Code and the data are available here\footnote{\url{https://github.com/tunazislam/latent-themes-llms}}.
\begin{table}[h]
\begin{center}
 \scalebox{0.99}{\begin{tabular}{>{\arraybackslash}m{1.3cm}|>{\arraybackslash}m{6.2cm}}
 \toprule
 \textsc{\textbf{Climate}} & Economy\_pro, ClimateSolution, Pragmatism, Patriotism, AgainstClimatePolicy, 
 Economy\_clean, FutureGeneration, Environmental, HumanHealth, Animals, SupportClimatePolicy, AltEnergy, PoliticalAffiliation.\\
 \hline
 \hline
 \textsc{\textbf{COVID-19}} & GovDistrust, GovTrust, VaccineRollout, VaccineSymptom, VaccineEquity, VaccineStatus, EncourageVaccination, VaccineMandate, VaccineReligion, VaccineEfficacy, VaccineDevelopment, CovidPlan, Vote, VaccineMisinformation, NaturalImmunity.\\
 \bottomrule
\end{tabular}}
\end{center}
\caption{Pre-existing themes from previous studies: Climate campaigns \cite{islam2023analysis}, COVID-19 vaccine campaigns \cite{islam2022understanding}.}
\label{tab:pre_thm}
\end{table}
\section{Related Work}
Several studies have been conducted to understand messaging on social media \cite{islam2023weakly, islam2023analysis, islam2022covidfbAd, capozzi2021clandestino, silva2021covid, silva2020facebook, serrano2020political, mejova2020covid, capozzi2020facebook, andreou2019measuring}.
Most of the approaches to understanding messaging have relied on broader categories. However, in this paper, we focus on gaining a deeper understanding of messaging. 

%
Previous works \cite{islam2023weakly,islam2023analysis,islam2022covidfbAd,capozzi2021clandestino} used a predefined set of labels, themes, and arguments to analyze messaging. These were fixed and established based on existing topics or theoretical frameworks. However, previous works often fall short of capturing the nuances of messaging choices. For instance, within the \textbf{Economy} topic, it may neglect to recognize conflicting arguments, such as Argument 1: ``\textit{Increasing the minimum wage boosts economic growth}," versus Argument 2: ``\textit{Increasing the minimum wage harms small businesses and reduces employment opportunities}."
Recently, \citet{pacheco2022holistic,pacheco2022interactively}
suggested a human-in-the-loop approach and \citet{pacheco2023interactive} developed an interactive clustering approach to uncover new themes. In contrast to their work, we use LLMs inference to shape the emergent clusters.

Recently, large language models (LLMs) have made significant progress in learning from prompts via in-context learning (ICL) \cite{chowdhery2023palm,kojima2022large,le2022bloom,brown2020language}.
Several studies indicate that LLMs exhibit better performance in tasks traditionally completed by humans \cite{gilardi2023chatgpt,de2023can,dai2023llm,chiang2023can,ziems2024can}, highlighting a potential to leverage LLMs effectively in our task.
Researchers in the fields of qualitative research (QR) and Natural Language Processing (NLP) are currently investigating how LLMs can be utilized for TA. \citet{dai2023llm} proposed LLM-in-the-loop approach for TA. \citet{xiao2023supporting} combined GPT-3 with expert-drafted codebooks for supporting qualitative analysis. \citet{gao2023collabcoder} suggested a collaborative interface for qualitative analysis in their work, utilizing LLMs for generating initial codes and aiding in decision-making processes. 
In contrast, our framework employs
LLMs-in-the-loop to uncover the latent themes in messaging. In this paper, we develop a new set of themes (or codes) specifically tailored for analyzing messaging in a
particular domain, which helps speed up the development of domain-specific labels.

\section{Machine-in-the-Loop Framework}
Our methodology is meticulously designed to discover themes in social media messaging through a robust three-step process. This approach not only capitalizes on advanced computational techniques but also integrates human insight, ensuring a comprehensive and accurate analysis.
Importantly, our framework is flexible regarding the choice of representation. Both the embedding technique used for generating representations and the scoring function can be readily substituted with alternatives.
\subsection{Theme and Instance Representation}
\label{ad_thm_rep}
We represent instances and explanatory phrases under each theme using the Sentence BERT \cite{reimers2019sentence} embedding. To evaluate the degree of relevance between an instance and a specific theme, we adopt a straightforward distance-based method. This involves calculating the cosine distance between the instance and every explanatory phrase and example related to the themes, selecting the smallest distance score as the representative value.  
Our framework is agnostic of the representation used. The underlying embedding objective, as well as the scoring function, can easily be replaced.
\subsubsection{Grounding}
Before assigning instances to themes, every instance is grouped into unnamed clusters, effectively leaving them unassigned. Upon the first invocation of this process, instances are assigned to the theme that is nearest to them, but only if the newly calculated distance is shorter than, or at most equal to, the distance from their prior assignment. Previous assignments can correspond either to different themes or to the unnamed space. Note that this way, some instances can remain unassigned based on a threshold.
\begin{figure}[h]
    \centering
	\includegraphics[width=1\columnwidth]{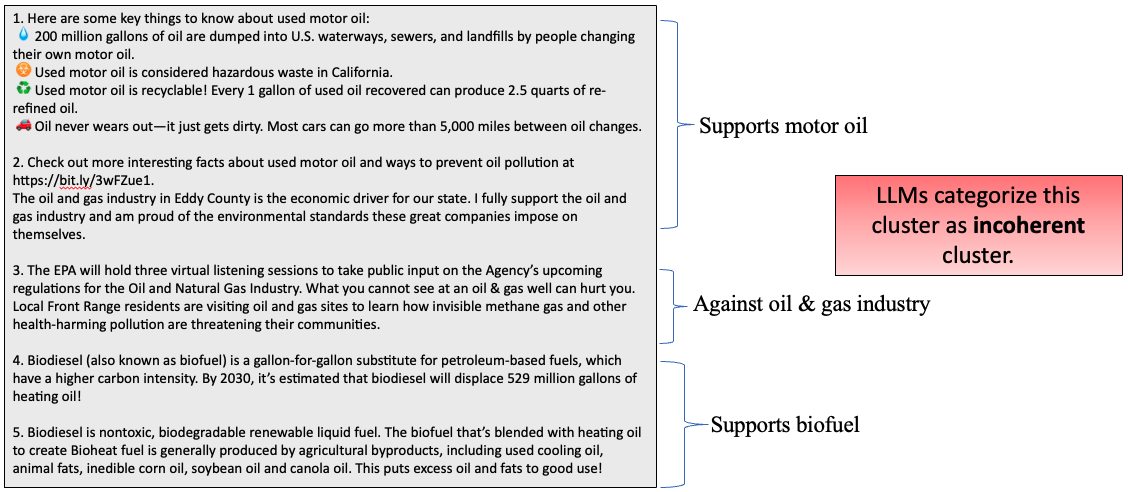}
	\caption{Example of an incoherent cluster in Climate.}
	\label{fig:incoh}
\end{figure}
\begin{figure}[h]
    \centering
	\includegraphics[width=1\columnwidth]{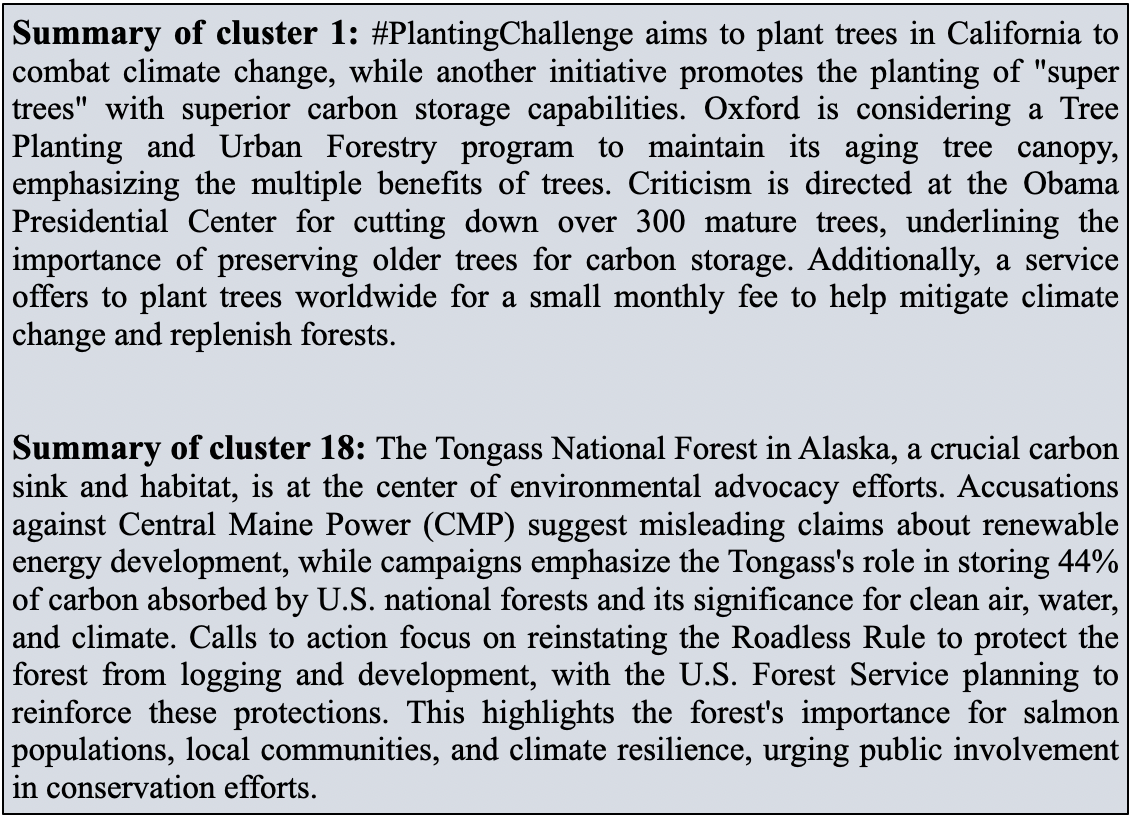}
	\caption{Example of a merged cluster in Climate.}
	\label{fig:merge}
\end{figure}
\subsection{Candidate Generation}
\label{cg}
The initial phase of our methodology, candidate generation, begins with the clustering of unassigned instances. We employ K-means clustering \cite{Jin2010}, a method chosen for its effectiveness in handling large datasets, which consist of varied and complex social media messaging. The K-means algorithm segments these instances into distinct clusters based on their thematic and linguistic similarities.
\begin{figure}[h]
    \centering
	\includegraphics[width=1\columnwidth]{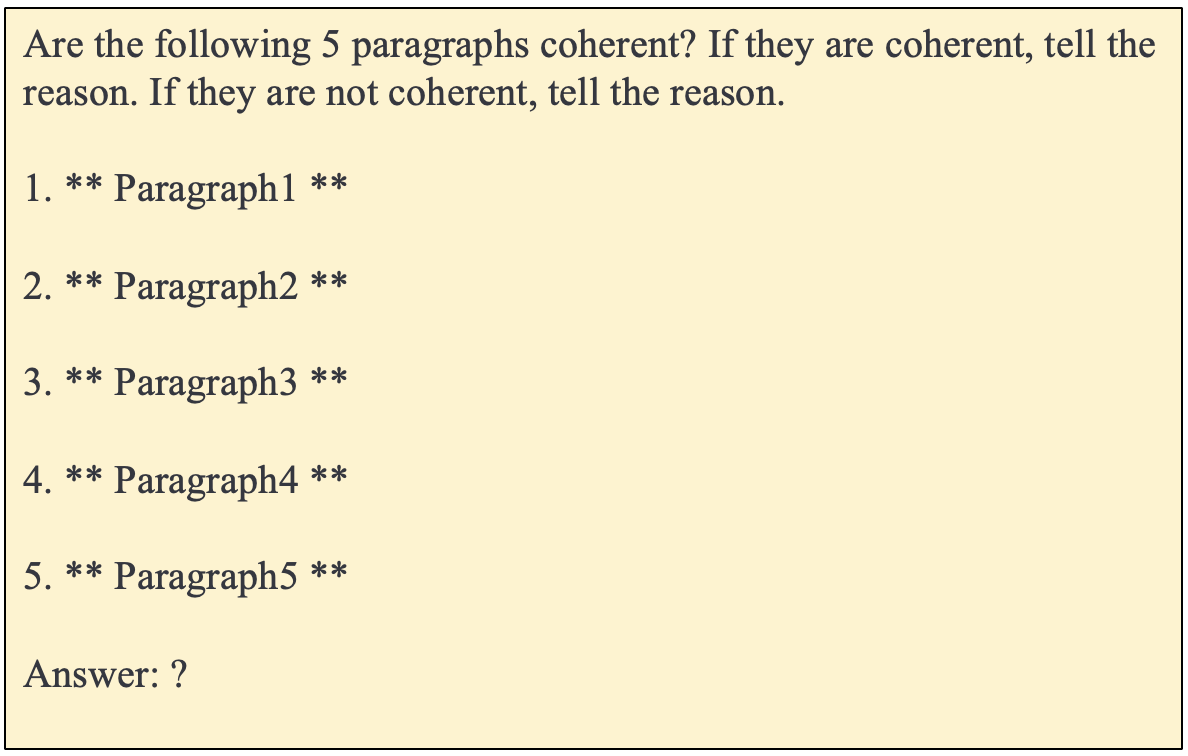}
	\caption{Prompt template for coherency check (shown as zero-shot).}
	\label{fig:coherency}
\end{figure}
\begin{figure}[h]
    \centering
	\includegraphics[width=1\columnwidth]{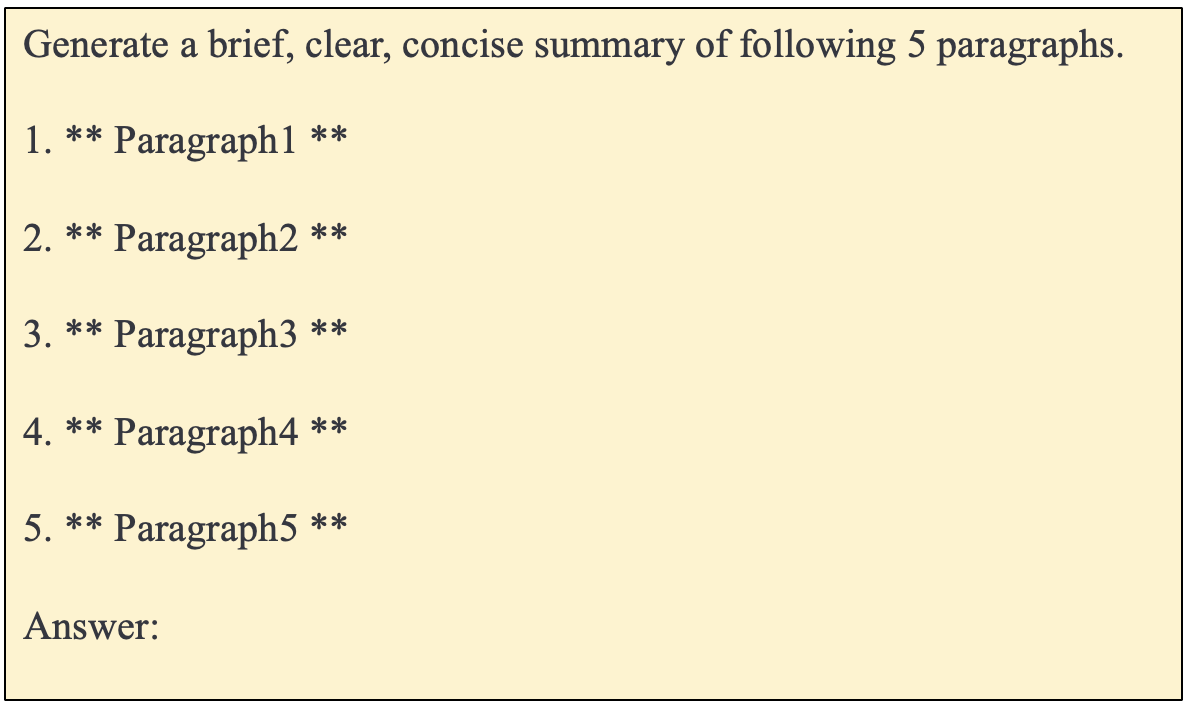}
	\caption{Prompt template for generating summary (shown as zero-shot).}
	\label{fig:summary}
\end{figure}
\begin{figure}[h]
    \centering
	\includegraphics[width=1\columnwidth]{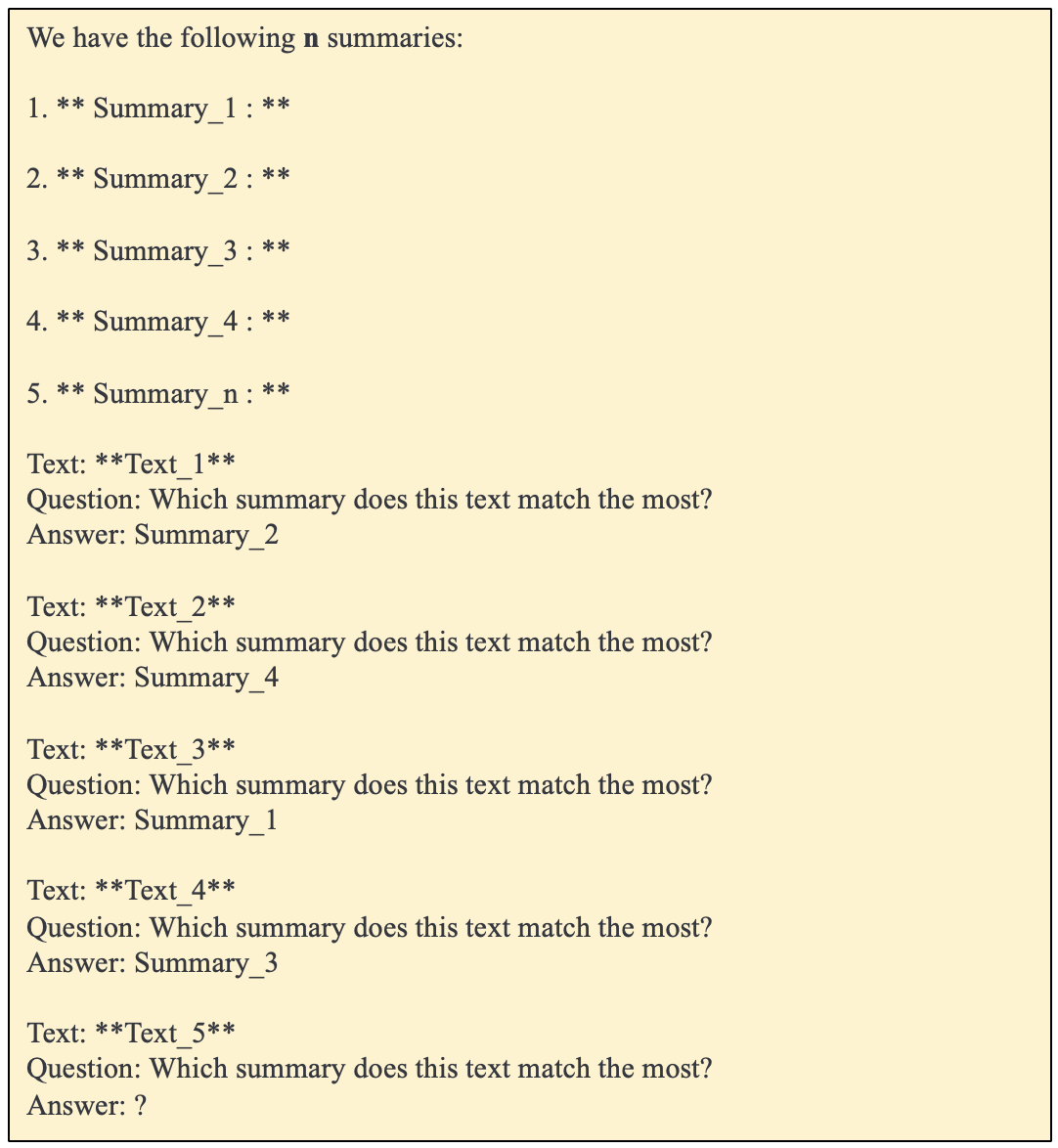}
	\caption{Prompt template for LLMs Mapper (shown as four-shot for climate).}
	\label{fig:map}
\end{figure}
\subsubsection{Coherency check}
To ensure the consistency and coherence of these clusters, we perform a cluster coherence check. We create subsets of instances from each cluster, with each subset containing $4$-$5$ instances. We then prompt LLMs in a zero-shot setting, asking if these instances belong together. Based on the responses provided by the LLMs, we identify incoherent clusters as noise and exclude them. It is important to note that we discard only the clusters, not the instances associated with them. These instances are still utilized for cluster membership assignments in the subsequent stage of our framework. Fig. \ref{fig:incoh} shows an example of an incoherent cluster from the dataset. 
These are the top $5$ ads that are closest to the centroid of a cluster. $1^{st}$ and $2^{nd}$ ads are about supporting motor oil, $3^{rd}$ one is against the oil and gas industry, $4^{th}$ and $5^{th}$ ads are about supporting biofuel. LLMs categorize this cluster as an \textit{incoherent} cluster (answering \textbf{RQ1}).
\subsubsection{Cluster Summarization}
To make the clusters interpretable, we prompt LLMs in a zero-shot setting to generate a multi-document summary based on top-$K$ instances of the clusters ($k=5$). 
\subsubsection{Redundancy check}
We embed the summaries using SBERT \cite{reimers2019sentence} and compute the cosine similarity score between the embedding summaries of cluster pairs. Based
on a threshold value ($\geq 0.6$), we \textit{merge} two clusters. 
Fig. \ref{fig:merge} shows an example of a merged cluster from the data. These cluster summaries are related to ``\textit{initiatives to plant new trees and protect existing forests to save planet}" and based on the threshold, we merge them. 
\subsubsection{Naming Cluster}
Finally, we prompt LLMs in a zero-shot manner to generate the short descriptive labels/titles for the newly discovered clusters.
\begin{table*}
    \centering
    \resizebox{1\textwidth}{!}{%
    \begin{tabular}{l|l|l|llll}
    \toprule
       \textbf{\textsc{Case}} & \multirow{2}{*}{\textbf{\textsc{Method}}} &  \textbf{\textsc{Num.}} & \multicolumn{4}{c}{\textbf{\textsc{Num. Covered Ads}}}\\
       \textbf{\textsc{Study}} & \textbf{\textsc{}} & \textbf{\textsc{Themes}} & \textsc{thr $< 0.6$} & \textsc{thr $< 0.5$} & \textsc{thr $< 0.4$} & \textsc{thr $< 0.3$} \\
    \midrule
     \multirow{3}{*}{Climate} &  Pre-existing & 13 & 14652 & 9725 & 3731 & 558 \\
    & +After Iter1 & 20 & 18702 & 14583 & 8646 & 2944\\
     & \textbf{+After Iter2} & \textbf{25} & \textbf{18988} & \textbf{15052} & \textbf{9079} & \textbf{3180} \\
     \midrule
      \midrule
   \multirow{3}{*}{COVID-19} & Pre-existing & 15 & 7889 & 6426 & 3480 & 771 \\
    & +After Iter1 & 20 & 8852 & 7627 & 4737 & 1302\\
    & \textbf{+After Iter2} & \textbf{23} & \textbf{9092} & \textbf{7898} & \textbf{5038} & \textbf{1590} \\
    \bottomrule 
    \end{tabular}}
    \caption{Coverage. THR: Threshold, NUM: Number.}
\label{table:cover}
\end{table*}
\begin{figure*}
\centering
\begin{subfigure}{1\columnwidth}
  \centering
  \includegraphics[width=\textwidth]{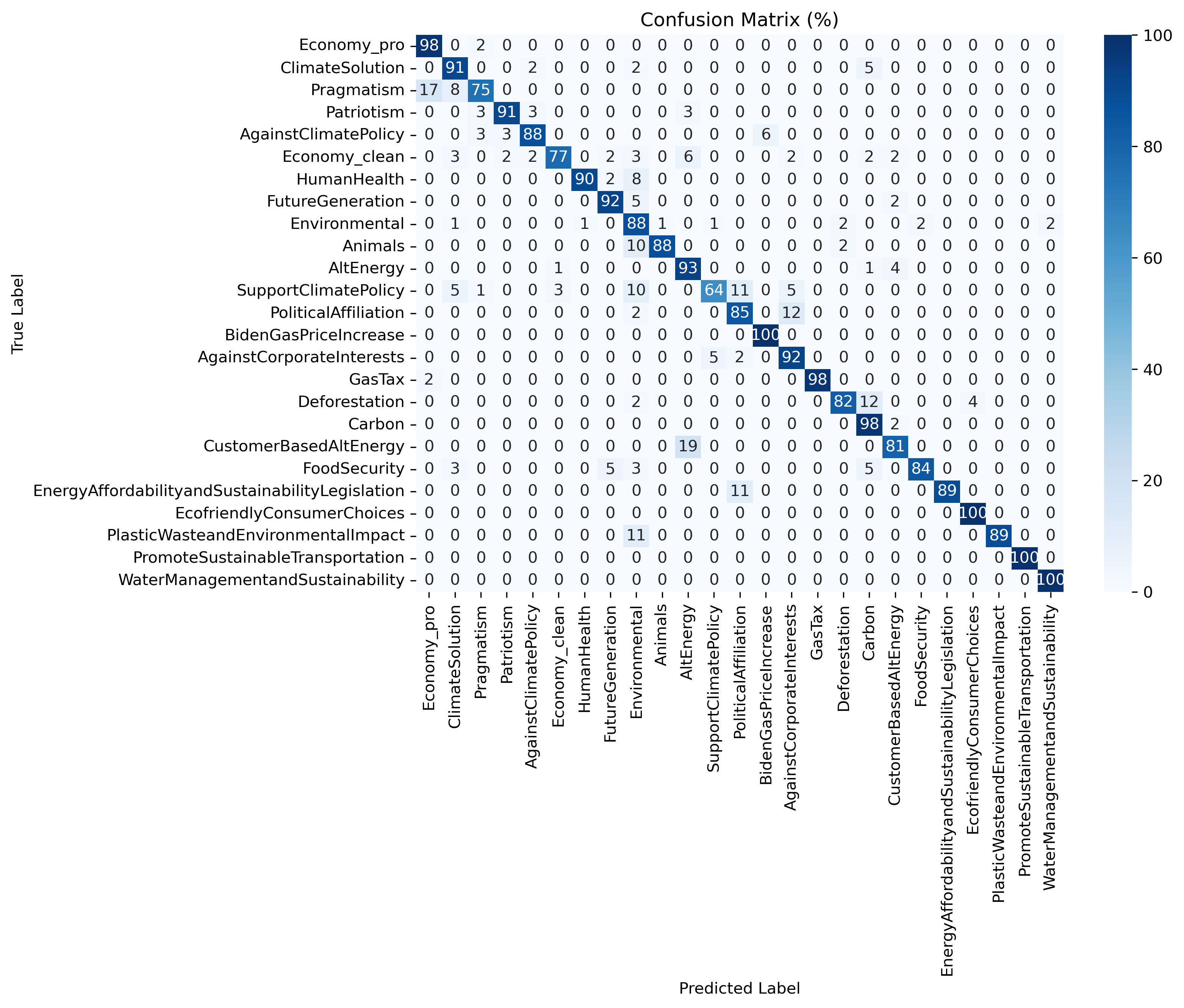}
  \caption{Assignment with LLMs mapper (Climate).}\label{fig:cm_llm}
\end{subfigure}%
\begin{subfigure}{1\columnwidth}
  \centering
  \includegraphics[width=\textwidth]{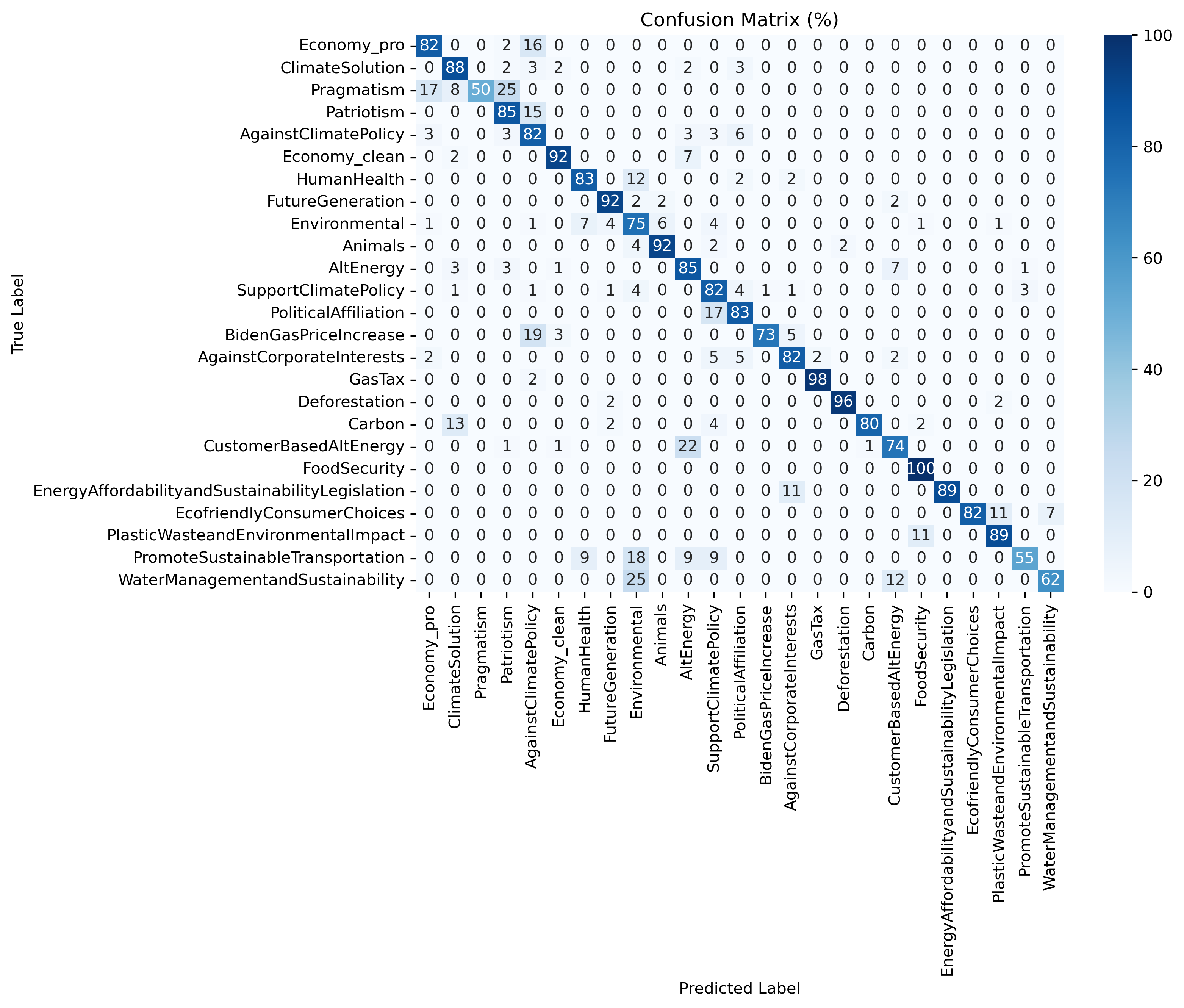}
  \caption{Assignment with SBERT (Climate).}\label{fig:cm_sbert}
\end{subfigure}
\begin{subfigure}{1\columnwidth}
  \centering
  \includegraphics[width=\textwidth]{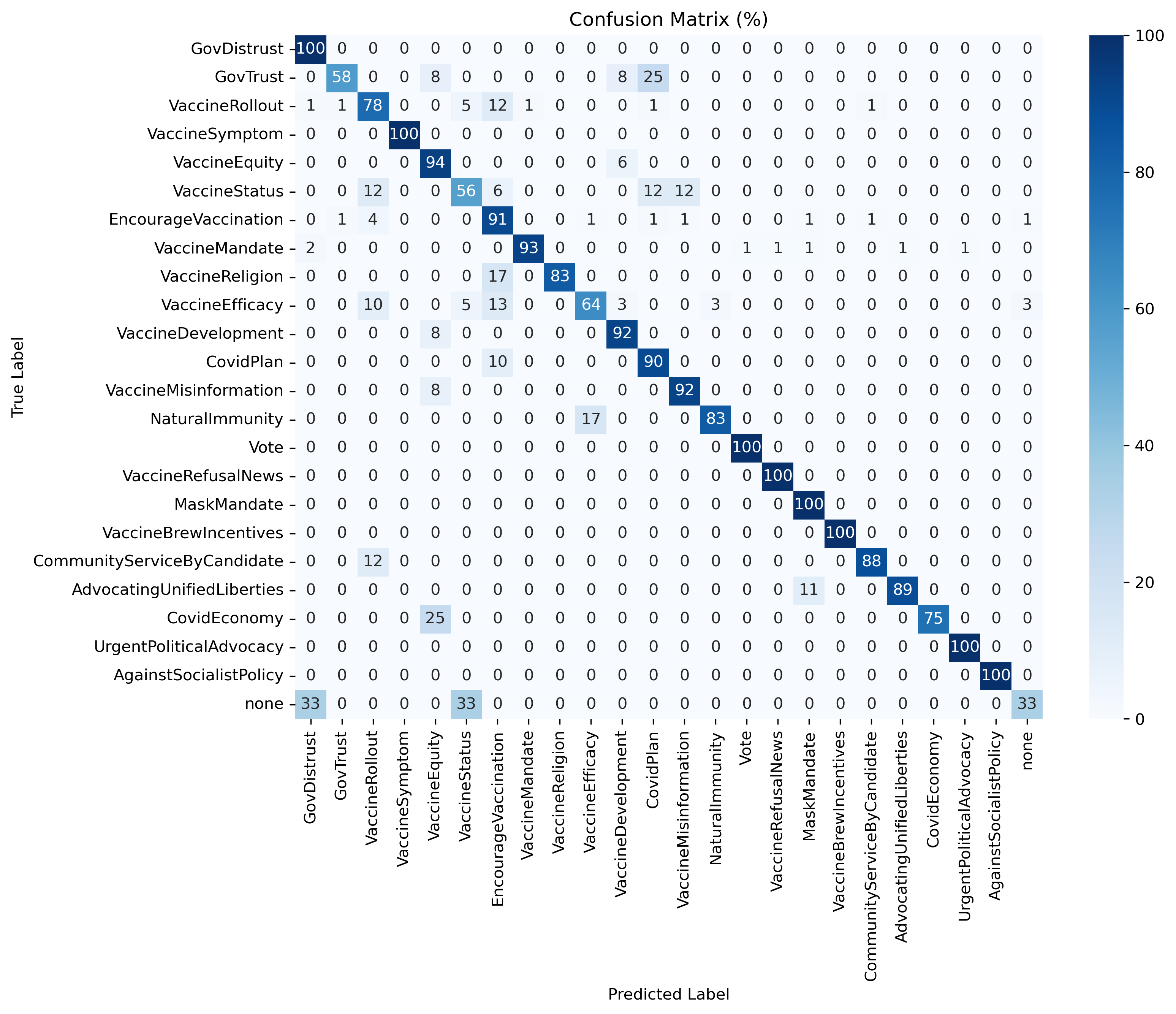}
  \caption{Assignment with LLMs mapper (COVID-19).}\label{fig:cm_llm_covid}
\end{subfigure}%
\begin{subfigure}{1\columnwidth}
  \centering
  \includegraphics[width=\textwidth]{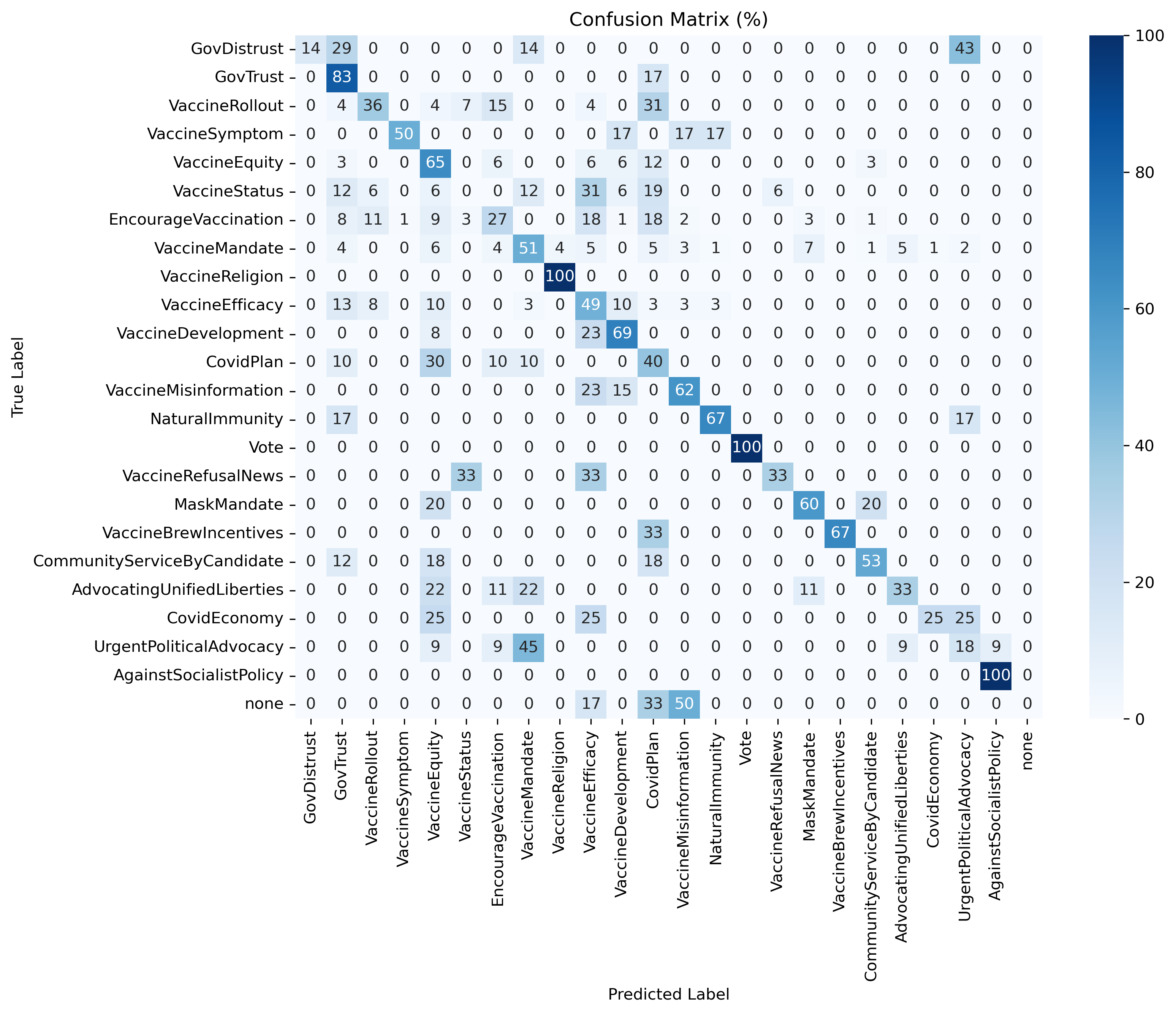}
  \caption{Assignment with SBERT (COVID-19).}\label{fig:cm_sbert_covid}
\end{subfigure}
\caption{Confusion matrices after $2^{nd}$ round of iteration + assignment for case studies climate and COVID-19.}
\label{fig:cm}
\end{figure*}
\subsection{Human Validation}
The second phase introduces an optional but crucial step – human validation. Two NLP and CSS researchers 
(age range $30$-$45$, $1$ male, and $1$ female) did one $2$-hour session to assess the clusters formed by the algorithms. The annotators included advanced graduate students and faculty. 
Their expertise is particularly invaluable in evaluating the nuances of language and context that automated systems might overlook. 
They scrutinize the merging decisions, the descriptiveness of the clusters, and overall coherence, bridging the gap between computational efficiency and the nuanced understanding of human judgment. 

For the \textit{cluster coherency} check, the annotators are
asked to verify if the top $5$ ads from a cluster
are coherent or not. They are asked
to provide a score of $1$ if the ads are coherent and a score of $0$ if they are not.

For the \textit{cluster summary} check, the annotators are
asked to verify if the generated summary is correct or not. They are asked
to provide a score of $1$ if the summaries are correct and a score of $0$ if they are not.
\subsection{Map Instances to Themes}
\label{mapper}
In the final phase, we provide summaries of all clusters (including pre-existing + newly discovered) and prompt LLMs in a few-shot setting to assess whether a new instance aligns with the summarized clusters. To alleviate the
issues of being hallucinated, we force LLMs to answer the questions by selecting from the given summarized clusters. 
The LLMs, equipped with the context from the initial phases, effectively categorize new texts into the relevant clusters (answering \textbf{RQ2}). This phase is crucial for maintaining the dynamism of the clusters, allowing the system to adapt to new data and emerging themes.
\section{Case Studies}
We investigate two case studies centered around discussions on social media: (1) climate campaigns and (2) COVID-19 vaccine campaigns. For climate campaigns, we work on the corpus of $21372$ ads and $13$ themes (Table \ref{tab:pre_thm}) released by \citet{islam2023analysis}. This dataset includes climate-related English ads on Facebook from the US, spanning from January $2021$ to January $2022$.
For COVID-19 vaccine campaigns, we use a corpus of $9920$ ads and $15$ themes (Table \ref{tab:pre_thm}) released by \citet{islam2022understanding} focusing on COVID-19 vaccine-related English ads on Facebook based on US from December $2020$ to January $2022$. 
Each ad in both corpora includes the following attributes: ad ID, title, ad description, ad body, funding entity, spend, impressions, and distribution of impressions broken down by gender (male, female, unknown), age (seven groups), and location down to the state level in the USA. Additional details about the datasets can be found in the original publications. 
\subsection{Results}
Initially, we assign pre-existing themes to the ads using SBERT. Pre-existing themes cover only a small portion of ads, as shown in the first row of Table \ref{table:cover} for climate campaigns and the fourth row of Table \ref{table:cover} for COVID-19 vaccine campaigns. We use a threshold determined by the distance from the cluster centroid to an ad, with shorter distances being preferable. For a threshold of less than $0.5$, there are $11,647$ ads from the climate campaigns and $3,494$ ads from the COVID-19 vaccine campaigns remaining unassigned, which we consider as noisy sets.
We then apply k-means clustering to those noisy sets, with $k = 29$ for climate and $k = 18$ for COVID-19. Following the candidate generation phase, we identify $7$ new themes for climate and $5$ new themes for COVID-19 vaccine campaigns after $1^{st}$ iteration. 
We use GPT-4 \cite{achiam2023gpt} for cluster coherency check, generating cluster summaries, and naming the clusters. 
In the second iteration, we repeat this process with $6,789$ climate ads and $2,293$ COVID-19 vaccine ads that remained unassigned (noisy sets) from the first iteration (again applying a threshold of $< 0.5$). This time, for k-means clustering on noisy sets, we choose $k = 20$ for climate and $k = 16$ for COVID-19. Iteration $2$ results in the discovery of $5$ additional themes for the climate campaigns and $3$ new themes for the COVID-19 vaccine campaigns.
The complete set of final themes from both case studies is presented in Table \ref{tab:new_thm} for each iteration. Summaries of the themes (clusters) for climate campaigns and COVID-19 vaccine campaigns are provided in the Appendix.

\subsection{Prompt Templates}
\label{pt}
This section shows the prompt templates used to check cluster coherency (Fig. \ref{fig:coherency}),
generate cluster summary (Fig. \ref{fig:summary}), and 
map ads to themes (Fig. \ref{fig:map}). An example for prompting GPT-4 using three-shot for the COVID-19 dataset is provided here\footnote{\url{https://platform.openai.com/playground/p/AVhxXcExPvQDWP16MT5SdV6u?mode=chat }}. 
\subsection{Coverage}
Our machine-in-the-loop framework successfully discovers new themes that cover a large portion of the messaging for both climate and COVID-19 vaccine campaigns, as detailed in Table \ref{table:cover}.
In the climate campaigns case study, with a threshold $< 0.4$, pre-existing themes initially cover only $17.5\%$ ads. After $1^{st}$ iteration of our machine-in-the-loop approach, we can cover $40.5\%$ (see the second row of Table \ref{table:cover}). By the end of the $2^{nd}$ iteration, we manage to cover $42.5\%$ climate ads (refer to the third row of Table \ref{table:cover}).
For the COVID-19 vaccine campaigns, with the same threshold, pre-existing themes can cover only $35.08\%$ ads. After the first iteration, we are able to cover $47.75\%$ (shown in the fifth row of Table \ref{table:cover}) and after $2^{nd}$ iteration, we further cover $50.79\%$ COVID-19 vaccine related ads (see the sixth row of Table \ref{table:cover}).
\begin{table}[t]
\begin{center}
 \scalebox{0.7}{\begin{tabular}{>{\arraybackslash}m{1.5cm}|> {\arraybackslash}m{0.7cm}|>{\arraybackslash}m{8.5 cm}}
\toprule
       \textbf{\textsc{Case}} & \multirow{2}{*}{\textbf{\textsc{Iter.}}} &  \multirow{2}{*}{\textbf{\textsc{Themes}}}\\
       \textbf{\textsc{Study}} & & \\
 \midrule
 \multirow{6}{*}{Climate} & 1 & BidenGasPriceIncrease, AgainstCorporateInterests, GasTax, Deforestation, Carbon, CustomerBasedAltEnergy, FoodSecurity. \\
 \cmidrule(r){2-3}
 & 2 & EnergyAffordabilityandSustainabilityLegislation, EcofriendlyConsumerChoices, PlasticWasteandEnvironmentalImpact, PromoteSustainableTransportation, WaterManagementandSustainability. \\
 \midrule
 \midrule
   \multirow{4}{*}{COVID-19} & 1 & VaccineRefusalNews, MaskMandate, VaccineBrewIncentives, CommunityServiceByCandidate, AdvocatingUnifiedLiberties. \\
 \cmidrule(r){2-3}
 & 2 & CovidEconomy, UrgentPoliticalAdvocacy, AgainstSocialistPolicy. \\
 \bottomrule
\end{tabular}}
\caption{Resulting themes after each iteration for climate and COVID-19 vaccine campaigns. Iter: Iteration.}
\label{tab:new_thm}
\end{center}
\end{table}
\begin{figure*}[h]
\centering
\begin{subfigure}{1\columnwidth}
  \centering
  \includegraphics[width=\textwidth]{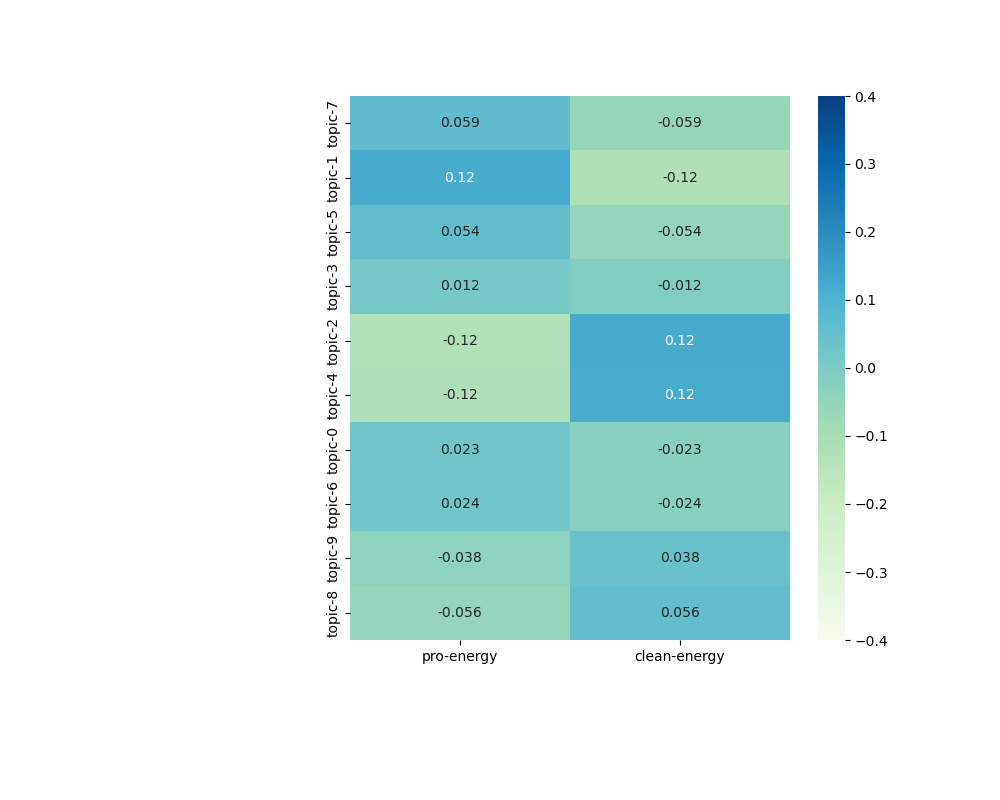}
  \caption{\textbf{Baseline}: 10 LDA Topics.}
  \label{fig:lda_10_climate}
\end{subfigure}%
\begin{subfigure}{1\columnwidth}
  \centering
  \includegraphics[width=\textwidth]{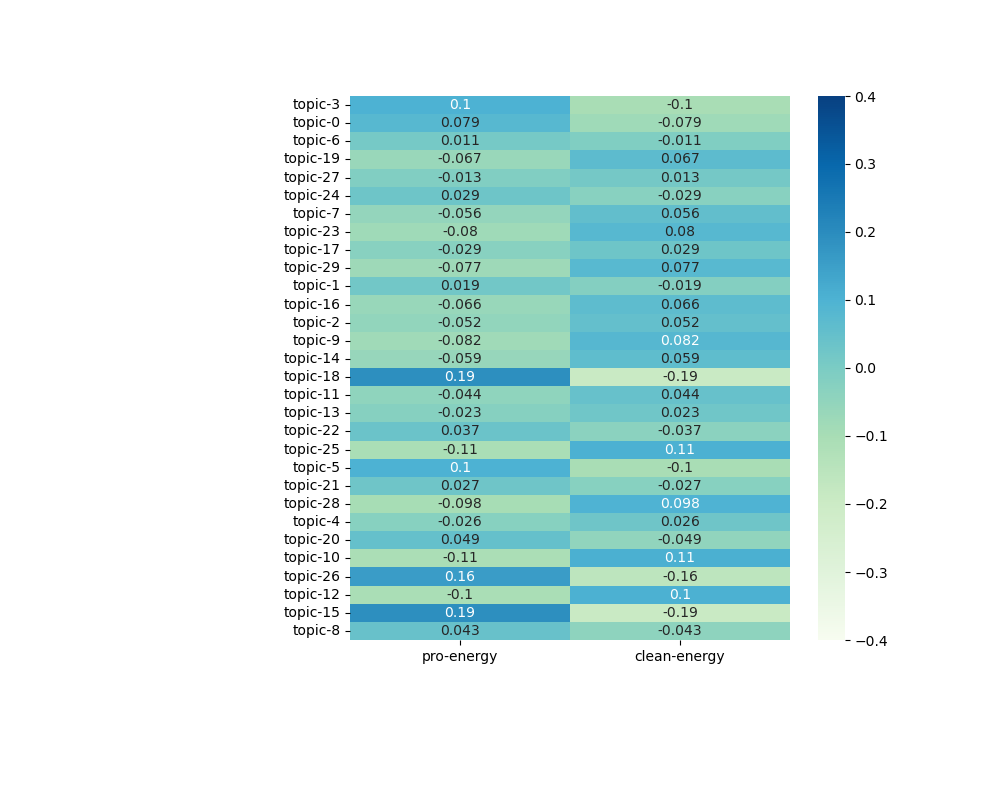}
  \caption{\textbf{Baseline}: 30 LDA Topics.}
  \label{fig:lda_30_climate}
\end{subfigure}
\begin{subfigure}{1\columnwidth}
  \centering
  \includegraphics[width=\textwidth]{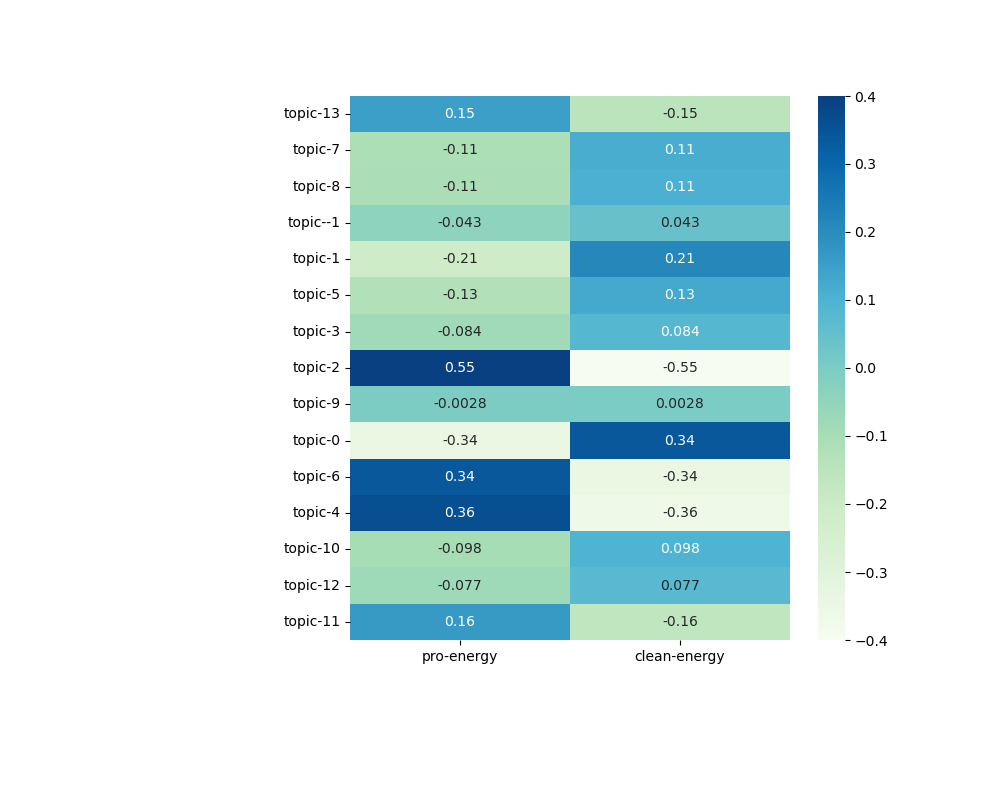}
  \caption{\textbf{Baseline}: 15 BERTopic Topics.}
  \label{fig:bertopic_15_climate}
\end{subfigure}%
\begin{subfigure}{1\columnwidth}
  \centering
  \includegraphics[width=\textwidth]{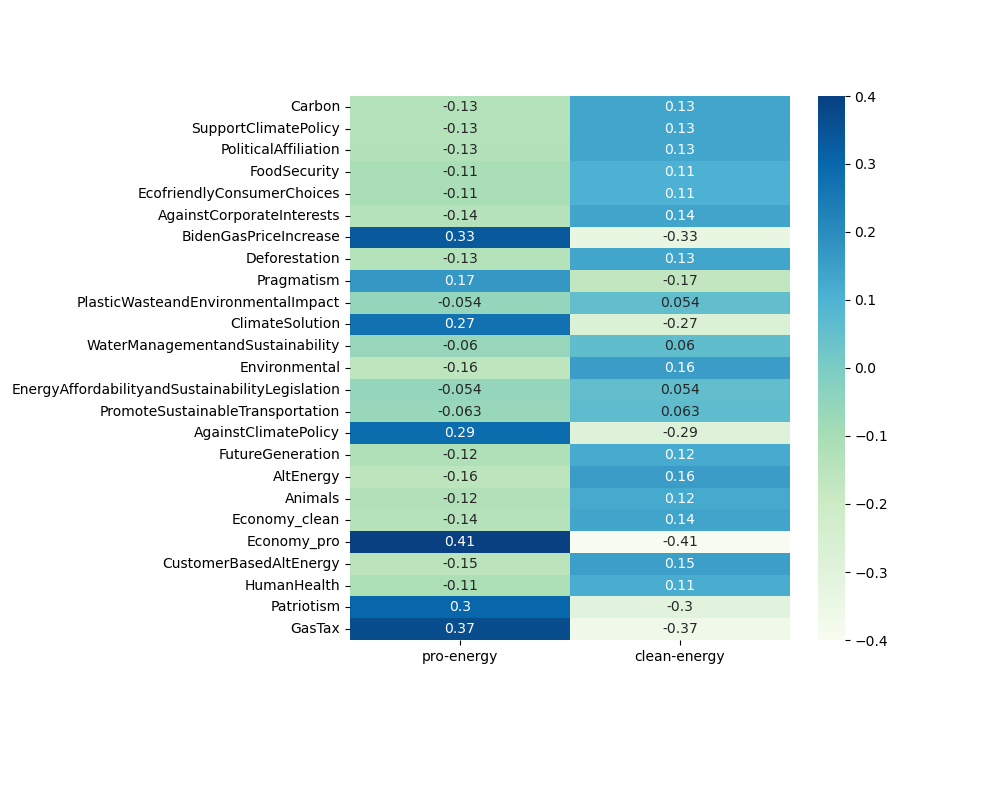}
  \caption{\textbf{Ours}: 25 themes after $2^{nd}$ round of iteration.}
  \label{fig:ours_climate}
\end{subfigure}
\caption{Correlations between \textbf{themes} and \textbf{stances} for \textbf{climate} case study.}
\label{fig:correl_climate}
\end{figure*}
\subsection{Mapping Quality}
To evaluate the \textit{mapping quality} of our model, we compare the performance of LLMs mapper with SBERT assignment (Baseline) for ads-to-themes mapping. We select ads that fall under threshold $< 0.4$ and then randomly sample $1072$ ads for climate campaigns. For COVID-19 vaccine campaigns, we randomly sample $565$ ads and we \textbf{do not} use a distance threshold in this case study. This approach is deliberately chosen to demonstrate the robustness of our methodology. Typically, ads that are closest to the cluster centroid in the embedding space exhibit higher accuracy and F1 scores for the SBERT assignment. For LLMs mapper assignment, we prompt LLMs in \textit{four-shot} settings for climate campaigns, and for COVID-19, we use a \textit{three-shot} prompting to assign ads to themes. However, our LLMs mapper assignment outperforms the SBERT assignment in both case studies, regardless of the distance selection. 
\begin{table}[h]
    \centering
    \resizebox{1\columnwidth}{!}{%
    \begin{tabular}{llcc}
        \toprule
        Case Study & \multirow{1}{*}{Method} & Acc. (\%) &  F1 (\%)  \\
        \midrule
        \multirow{2}{*}{Climate} & SBERT Assign. &  84.05 & 79.32  \\
        & \textbf{LLMs Mapper} & \textbf{88.15} & \textbf{89.24}  \\
        \midrule
        \midrule
         \multirow{2}{*}{COVID-19} & SBERT Assign. &  41.42 & 44.83  \\
        & \textbf{LLMs Mapper} & \textbf{85.49} & \textbf{81.74}  \\
        \bottomrule
    \end{tabular}}
    \caption{Mapping Quality w.r.t Human Judgements. Assign: Assignment, Acc: Accuracy, F1: Macro Average F1 score.}
    \label{tab:mapq}
\end{table} 
\begin{figure*}[h]
\centering
\begin{subfigure}{1\columnwidth}
  \centering
  \includegraphics[width=\textwidth]{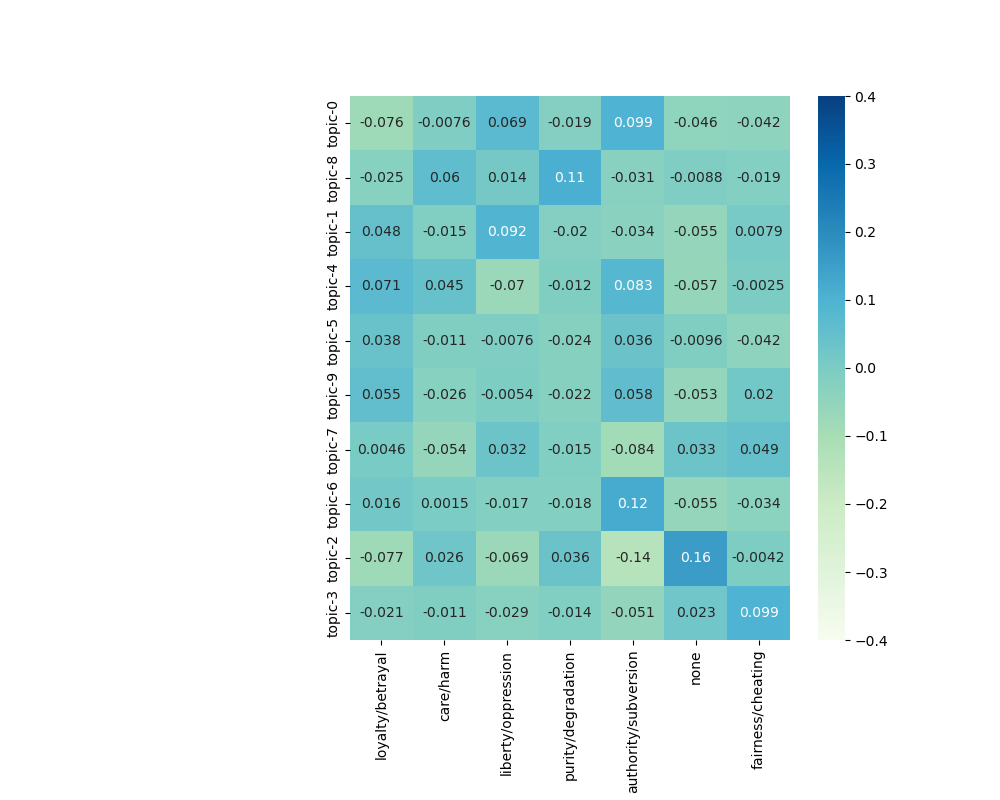}
  \caption{\textbf{Baseline}: 10 LDA Topics.}\label{fig:lda_10_covid}
\end{subfigure}%
\begin{subfigure}{1\columnwidth}
  \centering
  \includegraphics[width=\textwidth]{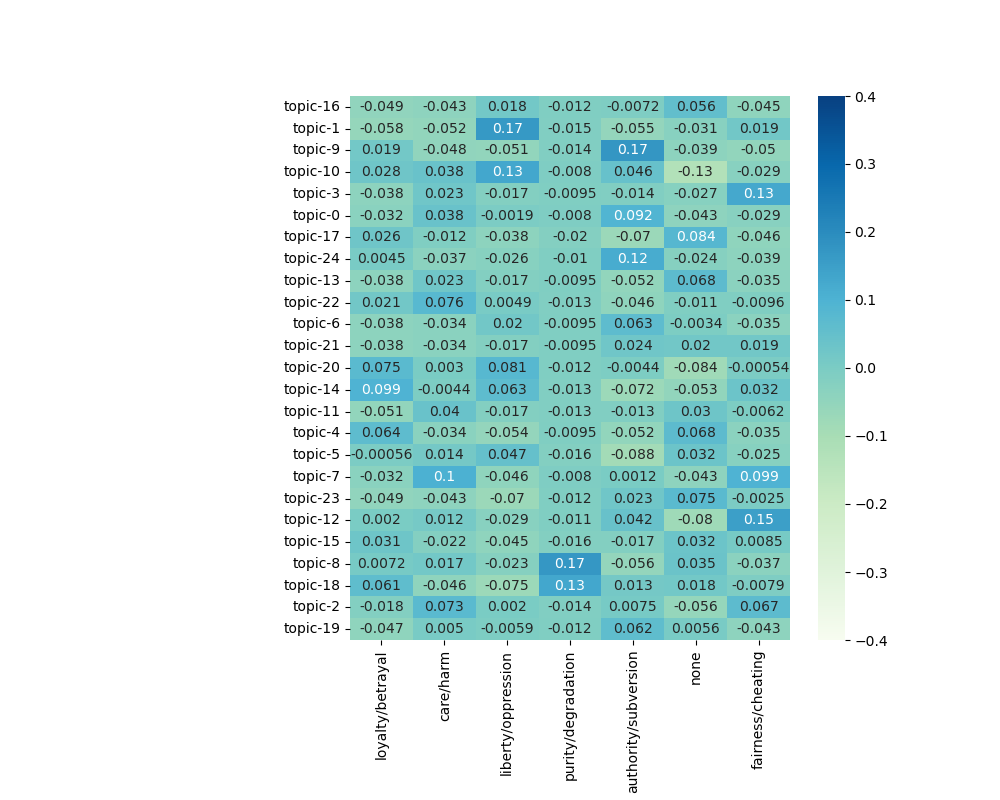}
  \caption{\textbf{Baseline}: 25 LDA Topics.}\label{fig:lda_25_covid}
\end{subfigure}
\begin{subfigure}{1\columnwidth}
  \centering
  \includegraphics[width=\textwidth]{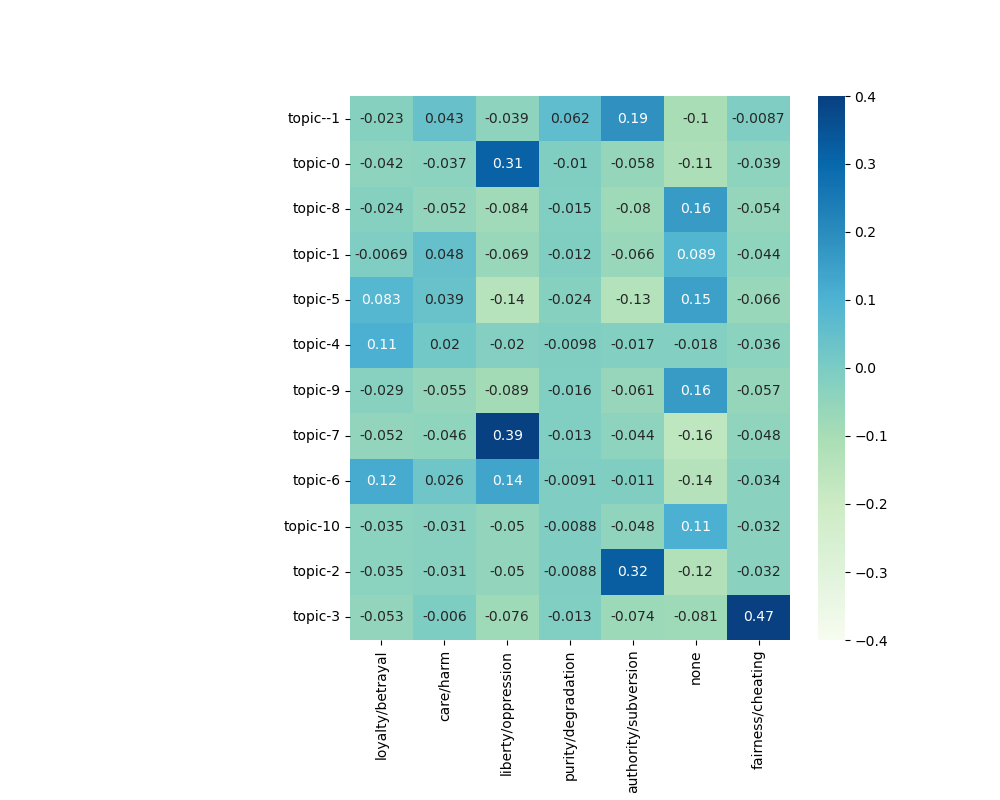}
  \caption{\textbf{Baseline}: 15 BERTopic Topics.}\label{fig:bertopic_15_covid}
\end{subfigure}%
\begin{subfigure}{1\columnwidth}
  \centering
  \includegraphics[width=\textwidth]{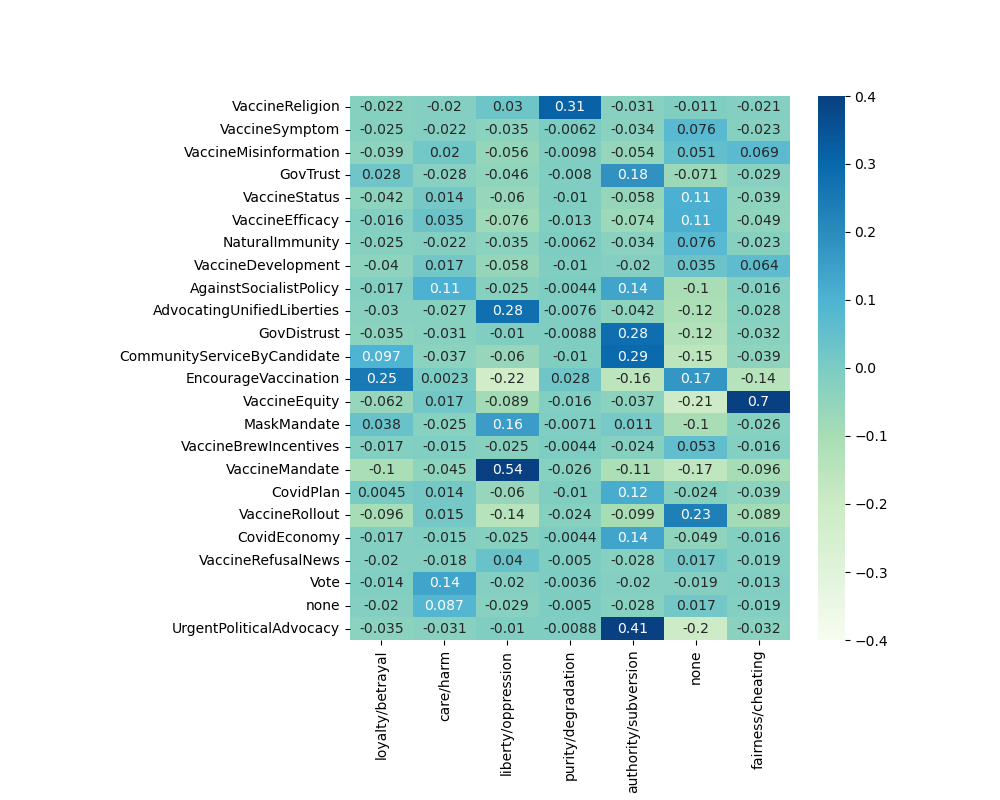}
  \caption{\textbf{Ours}: 23 themes after $2^{nd}$ round of iteration.}\label{fig:ours_covid}
\end{subfigure}
\caption{Correlations between \textbf{themes} and \textbf{moral foundations} for \textbf{COVID-19} case study.}
\label{fig:correl_covid}
\end{figure*}

We ask the annotators to provide a score of $1$ if the ad aligns with the theme,
$0$ for no alignment. There is no clear definition of alignment, and we let the
annotators make this judgment. 
Results are outlined in Table \ref{tab:mapq}.
The resulting mapping from ads-to-themes is fairly accurate (Fig. \ref{fig:cm_llm}) with respect to human judgments with an accuracy of $88.15\%$ and macro average F1 score of $89.24\%$ for the climate case study. On the other hand, we observe a noticeable improvement (Fig. \ref{fig:cm_llm_covid}) in accuracy ($85.49\%$) and F1 score ($81.74\%$) for COVID-19 case study.
Fig. \ref{fig:cm} shows the confusion matrix after $2^{nd}$ round of iteration of LLMs mapper and SBERT assignment. 
For climate, Fig. \ref{fig:cm_sbert} illustrates the SBERT mapper assignments, which are comparable to those of the LLMs mapper (Fig. \ref{fig:cm_llm}) due to ad selection from closest to the cluster centroid (threshold $< 0.4$). However, a significant difference is observed between the LLMs mapper (Fig. \ref{fig:cm_llm_covid}) and the SBERT mapper (Fig. \ref{fig:cm_sbert_covid}) in the COVID-19 case study, attributed to the random selection of ads independent of the distance threshold.
\begin{table*}
\centering
\begin{tabular}{p{1cm}|p{1cm}|c|p{2.5cm}|p{10cm}}
\hline
 Case Study & Age Group & \multirow{2}{*}{State} & \multirow{2}{*}{Theme}  & \multirow{2}{*}{Entity} \\
\hline
 \multirow{9}{*}{Climate} & \multirow{3}{*}{13-24} & \multirow{1}{*}{CA} & \multirow{1}{*}{Environmental} & Young People, Climate Change Denier, Climate Movement.   \\
 \cmidrule(r){3-5}
 & & \multirow{2}{*}{TX}  & \multirow{2}{*}{AltEnergy} & Natural Gas Systems, Coal Energy Systems, Nuclear Energy Systems, Wind Turbines, Solar Panels.\\
  \cmidrule(r){2-5}
 & \multirow{3}{*}{25-54 } & \multirow{1}{*}{CA} & \multirow{1}{*}{Carbon}  & Carbon Emissions, Sacramento Municipal Utility District.   \\
 \cmidrule(r){3-5}
 &  & \multirow{1}{*}{TX} & \multirow{1}{*}{AltEnergy} & Climate Change, Renewable Energy Plan.  \\
  \cmidrule(r){2-5}
 & \multirow{4}{*}{55+} & \multirow{1}{*}{CA} & \multirow{1}{*}{Animals} & Western Burrowing Owls, Aramis Industrial Solar Power Plant.  \\
 \cmidrule(r){3-5}
 & & \multirow{1}{*}{TX} & \multirow{1}{*}{SupporClimPolicy} & U.S. Senators or Congressman, Citizens' Climate Lobby.  \\
\hline
\hline
 \multirow{5}{*}{COVID} & \multirow{1}{*}{13-24} & \multirow{1}{*}{TX} & \multirow{1}{*}{VaccineRollout} & Kids/Children, Team/Teammates.   \\
 \cmidrule(r){2-5}
 \multirow{4}{*}{-19}& \multirow{1}{*}{25-54 } & \multirow{1}{*}{FL} & \multirow{1}{*}{UrgPoliticalAdvc}  & Ron DeSantis, Dr. Joseph Ladapo, Surgeon General, Floridians.   \\
  \cmidrule(r){2-5}
 & \multirow{2}{*}{55+} & \multirow{1}{*}{TX} & \multirow{1}{*}{EncourageVax} & Seniors, Pfizer, Elders.  \\
 \cmidrule(r){3-5}
 & & \multirow{1}{*}{FL} & \multirow{1}{*}{VaccineRollout} & Governor Ron DeSantis, Seniors, Bosters, Johnson \& Johnson vaccine.  \\
 \hline
\end{tabular}
\caption{Most mentioned entities, claim, and theme of targeted ads for \textbf{young}, \textbf{working-age}, \textbf{older} population of CA and TX from \textbf{climate} dataset. TX and FL from \textbf{COVID-19} dataset. SupClimPolicy: SupportClimatePolicy, UrgPoliticalAdvc: UrgentPoliticalAdvocacy, EncourageVax: EncourageVaccination.}
\label{tab:age}
\end{table*}
\vspace{-5pt}
\subsection{Qualitative Analysis}
We can characterize themes as the reasons people cite to support or oppose contentious topics, such as the climate change debate or the vaccine debate. We consider assignments to
be better if they are more cohesive, for example, when themes are more strongly correlated with specific stances or moral foundations.

For the climate debate case study, we perform a correlation test between the identified themes and the stance expressed in the ads (i.e., pro-energy or clean-energy). We calculate the Pearson correlation matrices and present them in Fig. \ref{fig:correl_climate}. We compare our discovered themes with a set of topics extracted using LDA. 
Additionally, we include one baseline, BERTopic \cite{grootendorst2022bertopic}, for this evaluation. BERTopic\footnote{\url{https://maartengr.github.io/BERTopic/#common}} 
is a hierarchical topic modeling and an off-the-shelf neural baseline for clustering texts. We observe that our themes (Fig. \ref{fig:ours_climate}) have stronger correlations with stances than the derived LDA topics (Fig. \ref{fig:lda_10_climate}, \ref{fig:lda_30_climate}). Although BERTopic (Fig. \ref{fig:bertopic_15_climate}) shows stronger correlations in some topics, our themes are more specific and cover a broader range of issues, from environmental to political themes. For example, we note that \textbf{pro-energy} stance is strongly correlated with themes such as \textit{gas price increase by Biden}, \textit{gas tax}, and \textit{patriotism} (Fig. \ref{fig:ours_climate}). 

For the COVID-19 vaccine debate case study, we calculate the Pearson correlation matrices to evaluate the correlation between themes and moral foundations, presenting the results in heatmaps (Fig. \ref{fig:correl_covid}).
We can interpret reasons as distributions over moral foundations. 
Our analysis reveals that our themes (Fig. \ref{fig:ours_covid}) exhibit higher correlations with moral foundations than those topics derived from LDA (Fig. \ref{fig:lda_10_covid}, \ref{fig:lda_25_covid}) and BERTopic (Fig. \ref{fig:bertopic_15_covid}).
For example, in Fig. \ref{fig:ours_covid}, we see that \textbf{liberty/oppression} moral foundation
is strongly correlated with reasons such as \textit{vaccine mandate}, \textit{advocating unified liberties}. The theme of \textit{urgent political advocacy} shows a higher correlation with the \textbf{authority/subversion} moral foundation. Additionally, the \textbf{purity/degradation} moral foundation is notably correlated with the theme of \textit{religious perspectives on taking or not taking the vaccine}. Other expected trends emerge, such as \textbf{fairness/cheating} being highly correlated with \textit{vaccine equity} (Fig. \ref{fig:ours_covid}). 
\subsection{Hyperparameters}
\label{hyper}
To obtain topics from LDA, we use Gensim \cite{rehurek2011gensim} implementation. We follow
the pre-processing steps shown in \citet{hoyle2021automated} and estimate the number of topics in a data-driven manner by maximizing. We do a grid search over a set of \{10, 20, 25, 30\} for the LDA baseline. For BERTopic, we choose the number of topics is $15$.

To determine the value of $k$ in k-Means, we follow both The Elbow Method and The Silhouette Method. For the climate case study, our optimal $k$ value is $29$ for iteration $1$ and $20$ for iteration $2$. On the other hand, our optimal $k$ value is $18$ and $16$ for iteration $1$ and $2$, respectively, for the COVID-19 case study.

For SBERT embedding, we use the sentence transformer model `all-mpnet-base-v2' with default parameters.
\begin{figure*}[t]
\centering
\begin{subfigure}{0.33\textwidth}
  \centering
  \includegraphics[width=\textwidth]{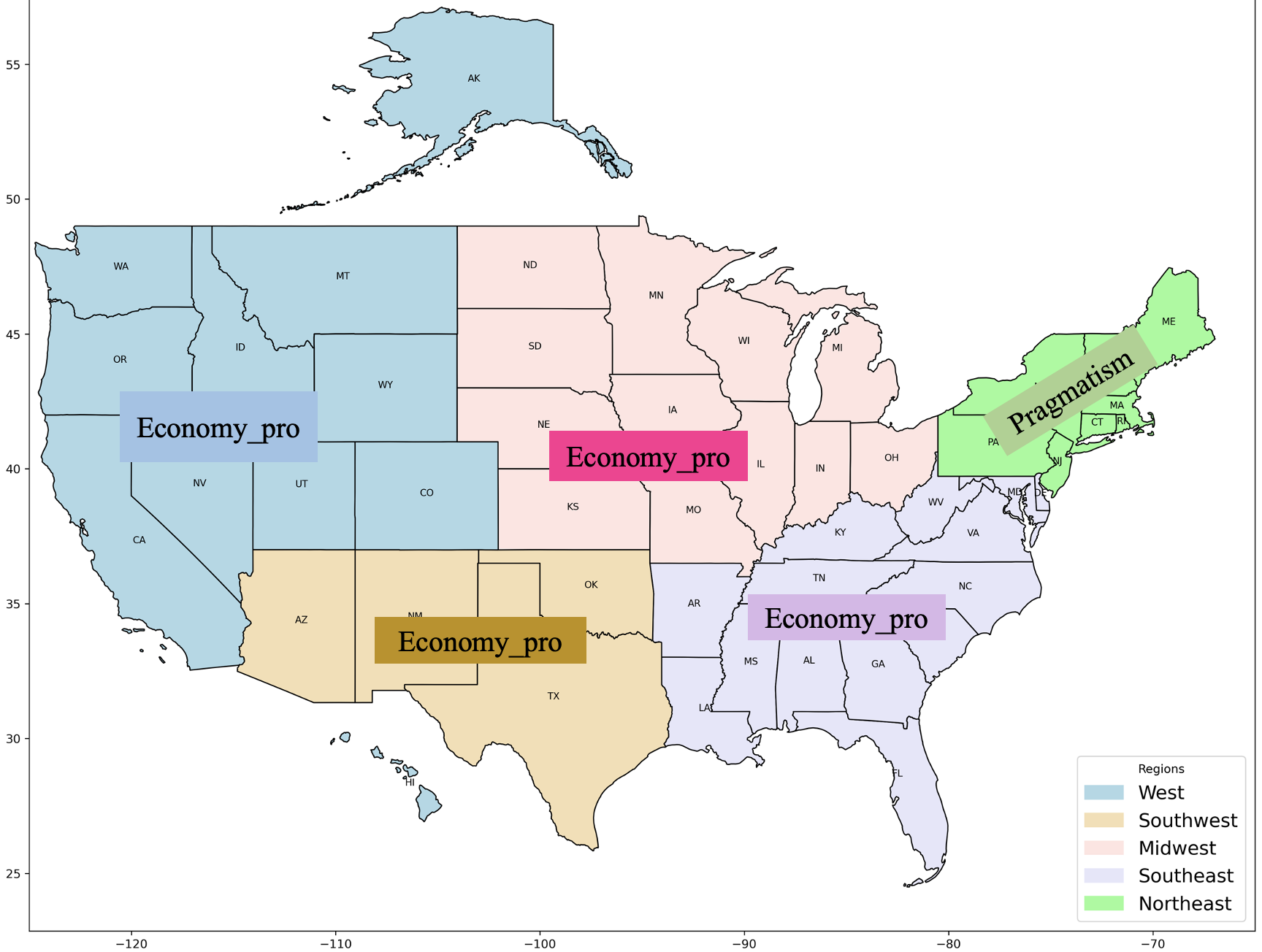}
  \caption{Exxon Mobil.}\label{fig:exn}
\end{subfigure}%
\begin{subfigure}{0.33\textwidth}
  \centering
  \includegraphics[width=\textwidth]{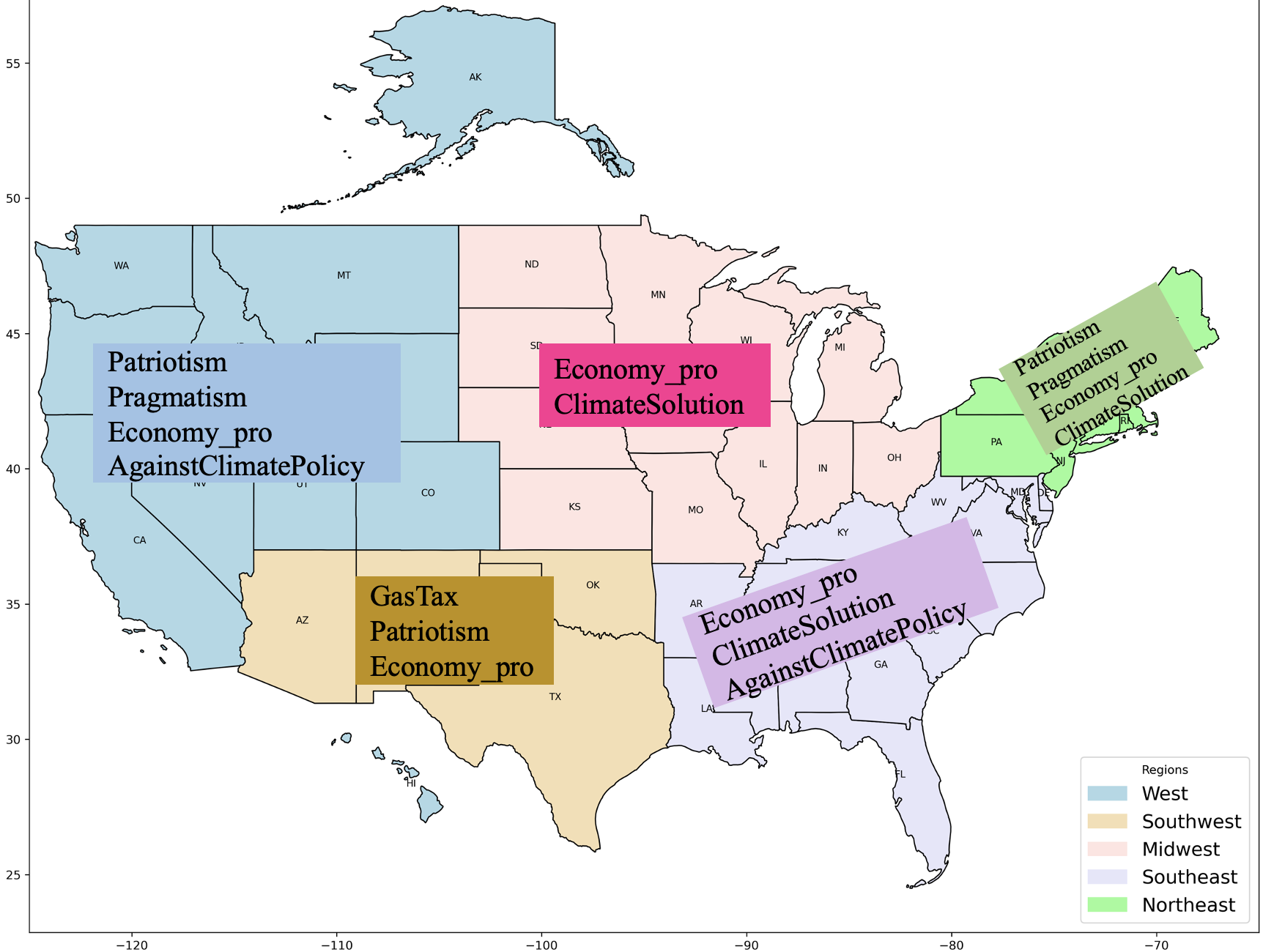}
  \caption{American Petroleum Institute.}
  \label{fig:pet}
\end{subfigure}%
\begin{subfigure}{0.33\textwidth}
  \centering
  \includegraphics[width=\textwidth]{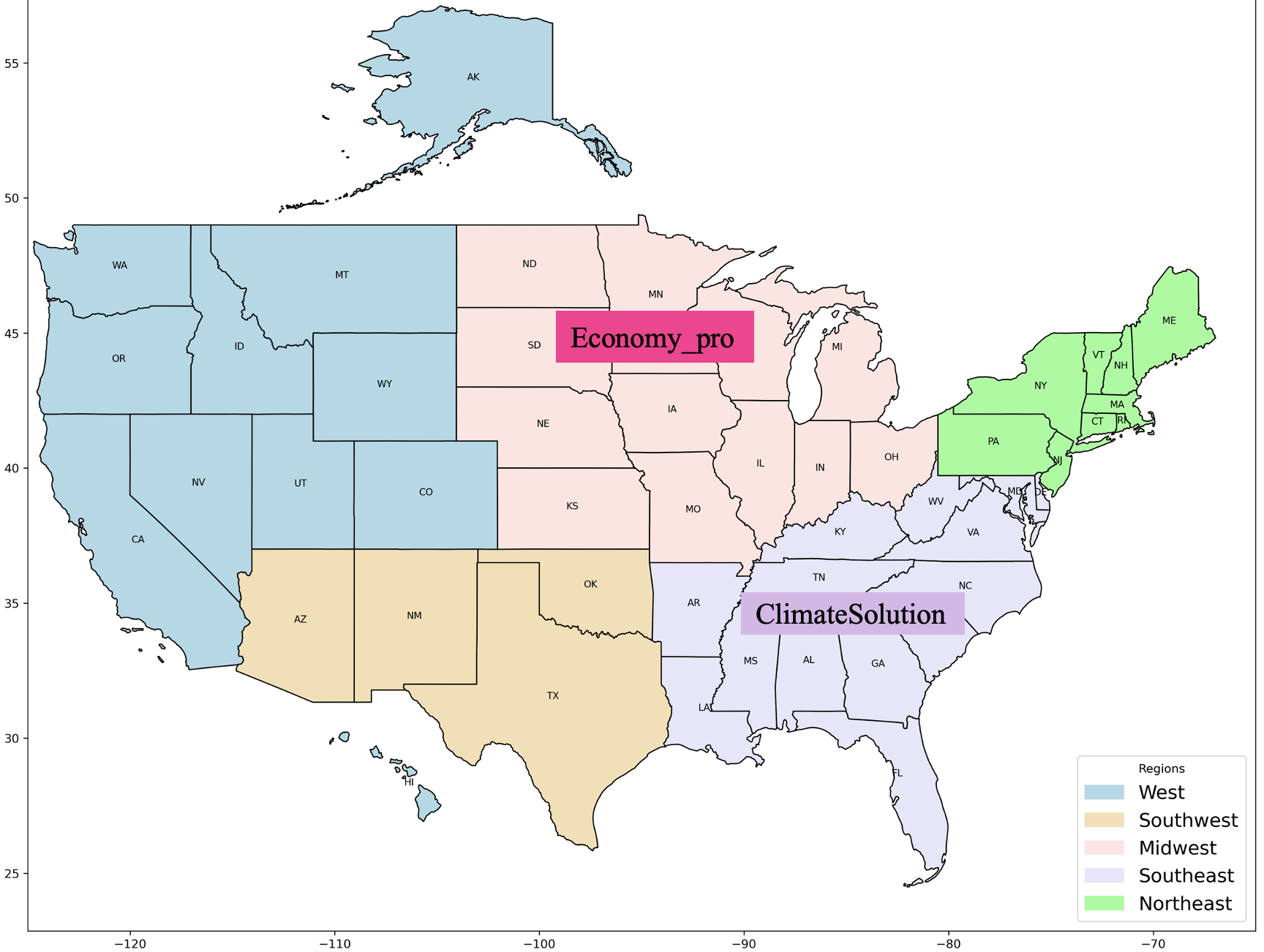}
  \caption{Consumer Energy Alliance.}\label{fig:cons}
\end{subfigure}
\begin{subfigure}{0.33\textwidth}
  \centering
  \includegraphics[width=\textwidth]{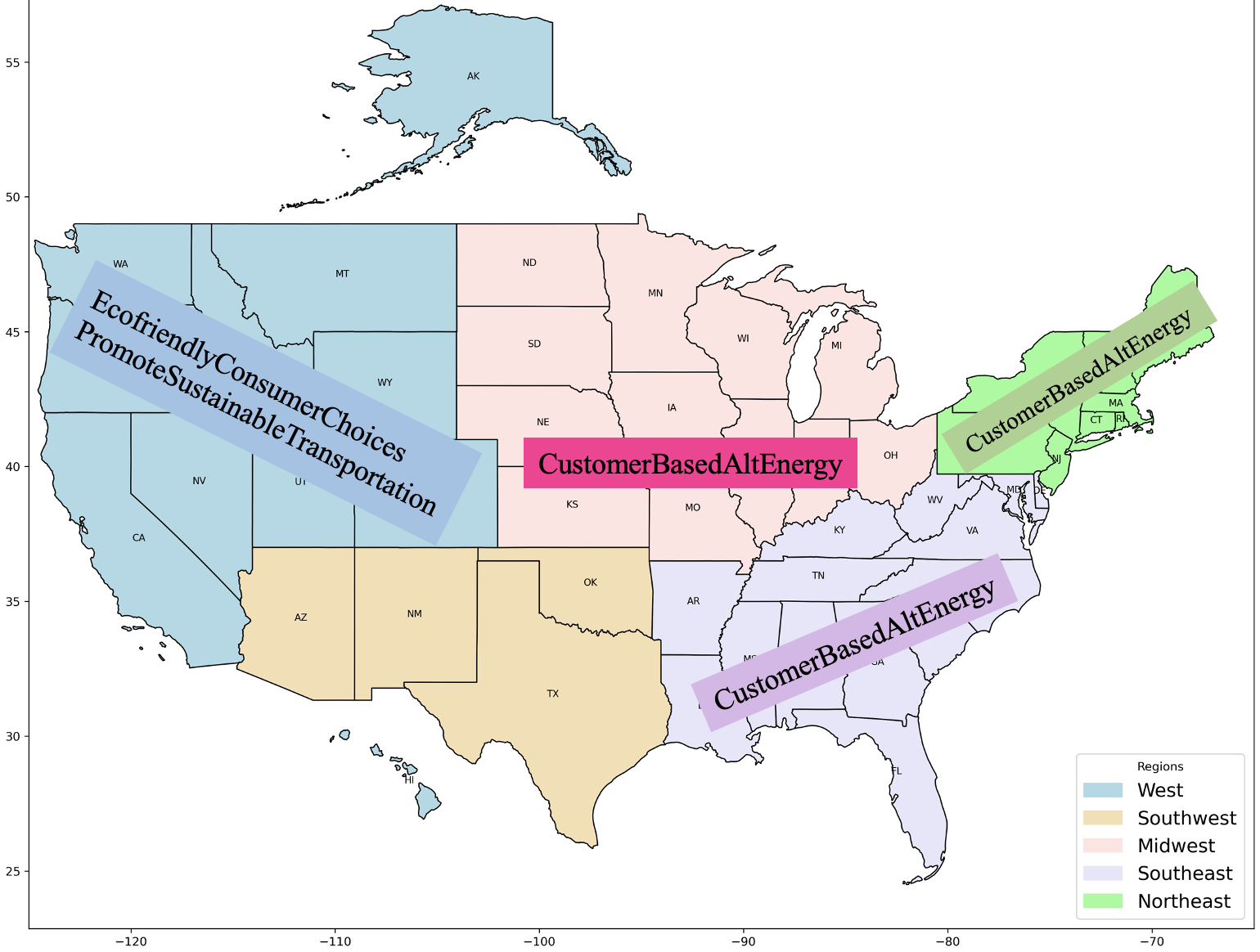}
  \caption{Apex clean energy.}\label{fig:apx}
\end{subfigure}%
\begin{subfigure}{0.33\textwidth}
  \centering
  \includegraphics[width=\textwidth]{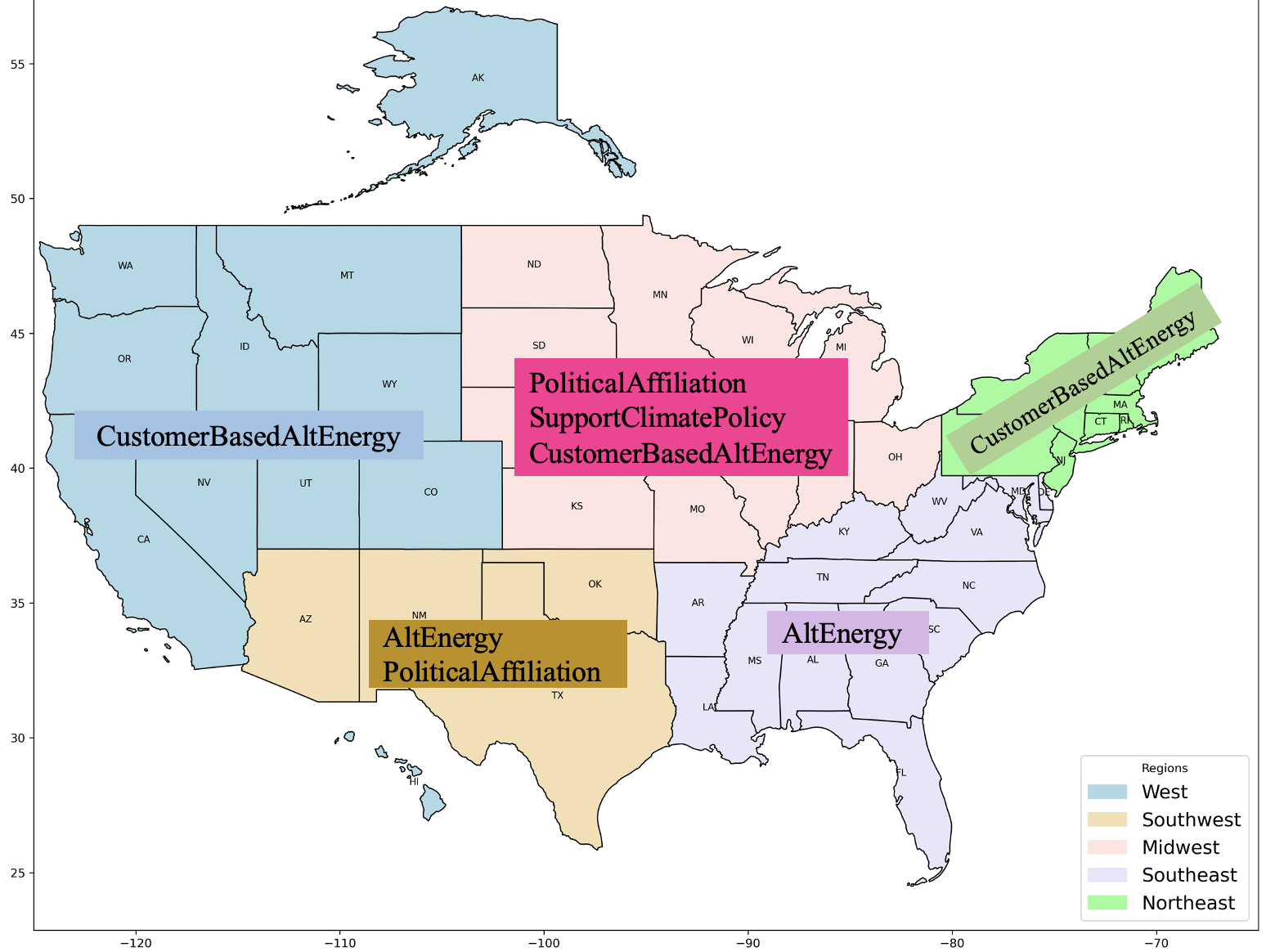}
  \caption{American Wind Energy Association.}
  \label{fig:wind}
\end{subfigure}%
\begin{subfigure}{0.33\textwidth}
  \centering
  \includegraphics[width=\textwidth]{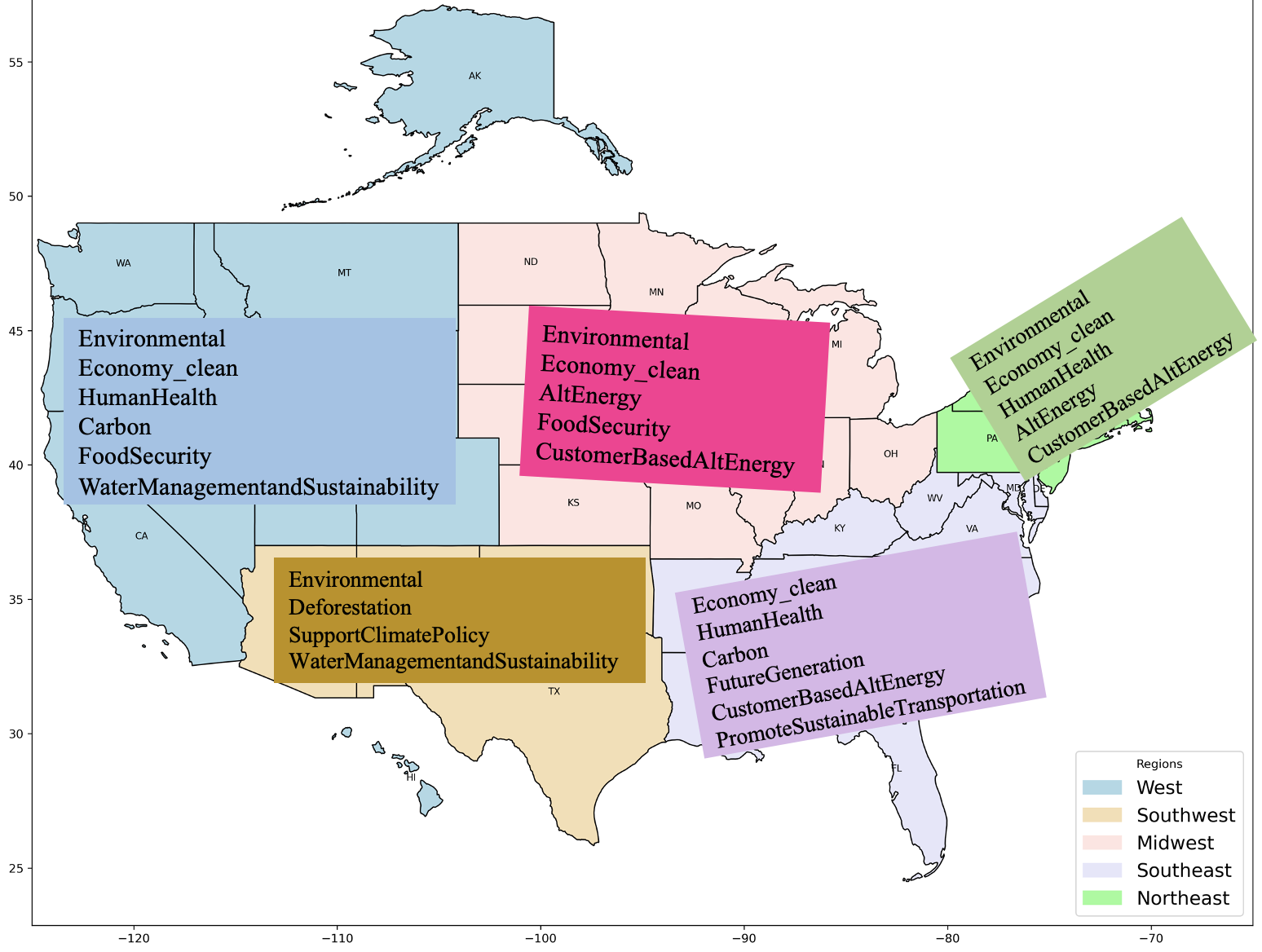}
  \caption{Climate Power Education Fund.}\label{fig:cp}
\end{subfigure}
\caption{Pro-energy and clean-energy Themes differ based on regions in the \textbf{climate campaigns} dataset. Funding entities are categorized based on Corporation, Industry Association, and Advocacy Group.}
\label{fig:geo_fe}
\end{figure*}
\begin{table}[h]
    \centering
    \begin{tabular}{p{1cm}|p{.85cm}|p{2cm}p{3cm}}
        \hline
        Case & \multirow{2}{*}{Gender} & \multirow{2}{*}{Theme} & \multirow{2}{*}{Entity} \\
        Study &  &  &  \\
        \hline
        \multirow{4}{*}{Climate} & \multirow{2}{*}{Male} & \multirow{2}{*}{GasTax} & Alex Askew, Karen Greenhalgh  \\
        \cmidrule(r){2-4}
        & \multirow{1}{*}{Female} & FutureGen & parents, children/kids  \\
        \hline
        \hline
        \multirow{10}{*}{COVID} & \multirow{4}{*}{Male} & \multirow{2}{*}{VaccineMandate} & Biden, federal employees.   \\
        \cmidrule(r){3-4}
         \multirow{7}{*}{-19} & & \multirow{2}{*}{VaccineEfficacy} & doctors, medical experts.  \\
        \cmidrule(r){2-4}
        & \multirow{8}{*}{Female} & \multirow{2}{*}{VaccineEfficacy} & pregnant, pregnancy, breastfeeding.   \\
        \cmidrule(r){3-4}
       & & \multirow{2}{*}{EncourageVax} & family, children/kids, community, friends.  \\
       \cmidrule(r){3-4}
       &  & \multirow{3}{*}{VaxMisinfo} & aborted fetal cells, microchips, mercury, mRNA, formaldehyde.  \\
       \hline
    \end{tabular}
    \caption{Mentioned entities and themes of targeted ads for \textbf{male only} and \textbf{female only} population from \textbf{climate campaigns} and \textbf{COVID-19 vaccine campaigns} datasets. FutureGen: FutureGeneration, EncourageVax: EncourageVaccination, VaxMisinfo: VaccineMisinformation.}
    \label{tab:gen}
\end{table}
\vspace{-5pt}
\section{Demographic Targeting}
Analyzing demographic targeting in social media messaging is vital for comprehensively understanding how messaging is tailored and disseminated to influence specific demographic groups.
\begin{figure*}[t]
\centering
\begin{subfigure}{0.5\textwidth}
  \centering
  \includegraphics[width=\textwidth]{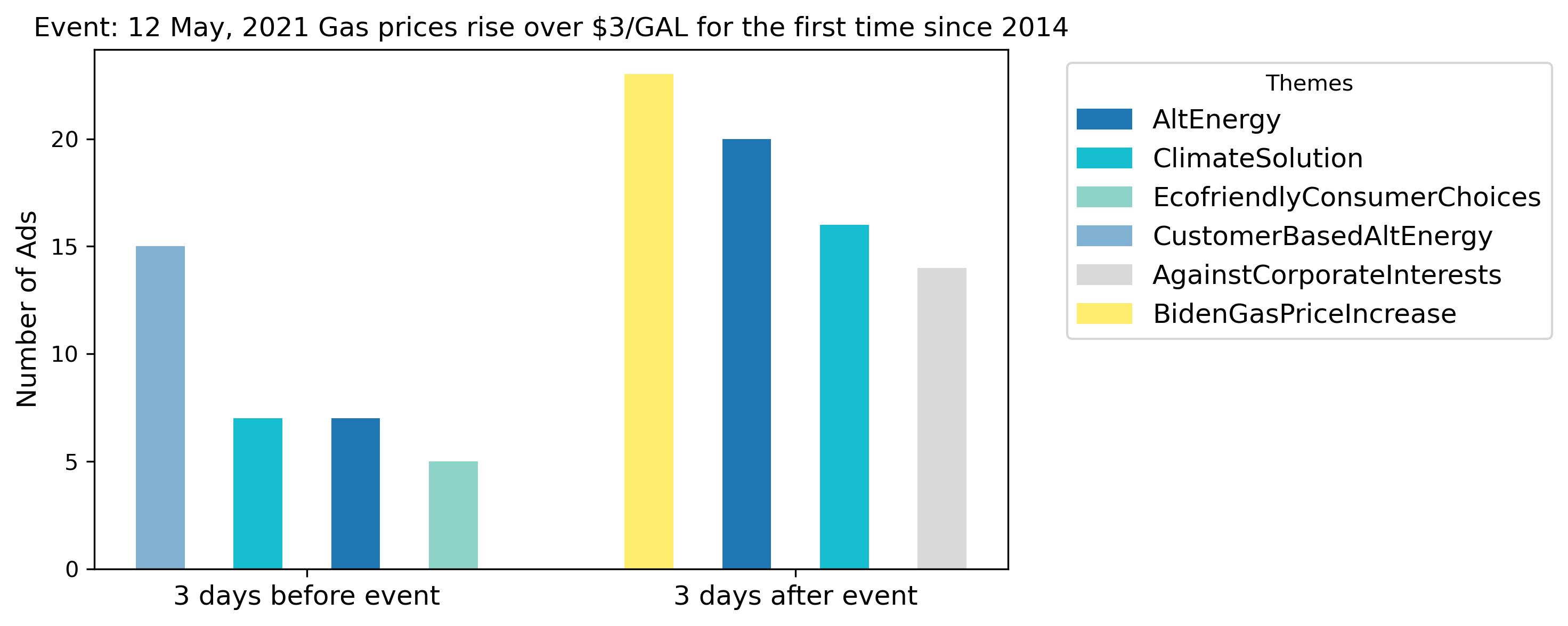}
  \caption{\textbf{Event:} Gas price increase on $12^{th}$ May $2021$.}
  \label{fig:gp}
\end{subfigure}%
\begin{subfigure}{.5\textwidth}
  \centering
  \includegraphics[width=\textwidth]{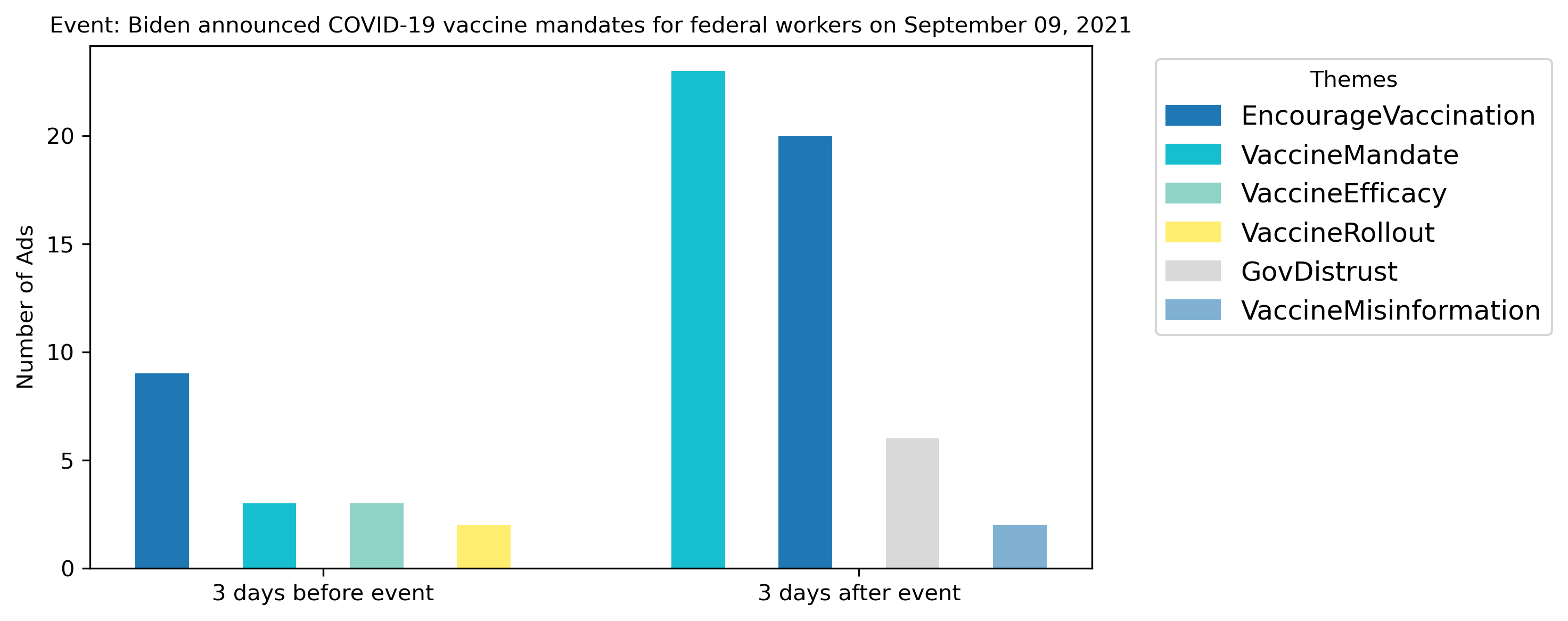}
  \caption{\textbf{Event:} Federal vaccine mandate on $9^{th}$ September $2021$.}
  \label{fig:vax}
\end{subfigure}
\caption{Advertisers' use of the top-$4$ themes before and after the events. (a) Themes `\textbf{BidenGasPriceIncrease}' and `\textbf{AgainstCorporateInterests}' emerged mainly after the occurrence of the event \textbf{Gas Price Increase}. (b) Advertisements related to the `\textbf{VaccineMandate}' theme increased after the \textbf{Federal Vaccine Mandate} announcement. Additionally, new themes such as `\textbf{GovDistrust}' and `\textbf{VaccineMisinformation}' have appeared following this mandate.}
\label{fig:event}
\end{figure*}
\subsection{How Themes Differ based on Age Group}
In the datasets (released by \citet{islam2023analysis,islam2022understanding}), there are $7$ age groups. In this paper, we categorize these into three distinct age groups: \textbf{young people} (ages $13$-$24$), \textbf{working-age people} (ages $25$-$54$), and the \textbf{older population} (age $55+$). We analyze the targeted ads for these $3$ age categories, focusing on the states of California (CA) and Texas (TX). The reason for choosing these two states is that TX is known as the energy capital of the world\footnote{\url{https://www.eia.gov/todayinenergy/detail.php?id=49356}}. Recently, CA has emerged as one of the most vocal states in the fight against climate change\footnote{\url{https://www.pewtrusts.org/en/research-and-analysis/blogs/stateline/2022/10/06/california-takes-leading-edge-on-climate-laws-others-could-follow}}. To identify the top-k most frequently mentioned entities in the texts, we prompt LLMs in a zero-shot setting. Similarly to extract the most mentioned claim, zero-shot prompting of LLMs has been used. 

Table \ref{tab:age} summarizes the results. Ads targeted for \textbf{young} population of CA focus on `\textbf{Environmental}' theme having claim `\textit{Importance of action against climate change to clean air and water}'. For the \textbf{working-age} demographic in CA, the ads focus on the `\textbf{Carbon}' theme, making claims like `\textit{The Sacramento Municipal Utility District (SMUD) aims to eliminate all carbon emissions from their power supply by 2030, contributing to the \#CleanPowerCity movement}'. In TX, both \textbf{young} and \textbf{working-age} groups are the focus of ads under the `\textbf{AltEnergy}' theme, each with specific claims. Interestingly, the \textbf{older} population in CA is targeted by ads with claims opposing the construction of solar plants, such as \textit{The construction of solar plants will harm wildlife and habitats}', aligning with the `\textbf{Animals}' theme. This approach is interesting, as it positions advertisers against solar plant construction while still engaging with the broader context of \textit{clean-energy} theme.

We adopt a similar approach to analyze themes across three different age groups in COVID-19 vaccine campaigns. We select Florida (FL) because the coronavirus surged to record levels there in August $2021$\footnote{\url{https://en.wikipedia.org/wiki/COVID-19_pandemic_in_Florida}}, and Texas (TX) because, as of April $3^{rd}$, $2021$, its vaccination rates were lagging behind the US average\footnote{\url{https://en.wikipedia.org/wiki/COVID-19_pandemic_in_Texas}}. Results are detailed in Table \ref{tab:age}.
For the \textbf{working age} group in FL, the ad's theme is `\textbf{UrgentPoliticalAdvocacy}' criticizing FL Governor Ron DeSantis's appointment of Dr. Joseph Ladapo as Surgeon General, characterizing Ladapo as an anti-masker and vaccine skeptic. While targeting an \textbf{older} population of TX, the theme is `\textbf{EncourageVaccination}' within seniors. Conversely, while targeting a \textbf{young} population of TX, the theme is `\textbf{VaccineRollout}', focusing on the vaccination of children to be a team player.
\vspace{-6pt}
\subsection{How Themes Differ based on Gender }
\vspace{-1pt}
To examine how messaging is customized for a particular gender, we pull advertisements that are specifically directed at either males or females.
For climate campaigns, we focus on the themes that are uncommon between these categories. We find that ads having `\textbf{GasTax}' theme focus only \textbf{male} population. On the other hand, `\textbf{FutureGeneration}' themed ads only target \textbf{female} population with mentioned entities `\textbf{\textit{Parents}}', `\textbf{\textit{Children/Kids}}' and the claim emphasizing the `\textit{Urgency of taking action against climate change to protect children's future}'. 
%

For COVID-19 vaccine campaigns, we focus on ads targeted exclusively towards males and females (Table \ref{tab:gen}). 
We find that ads have a `\textbf {VaccineEfficacy}' theme while focusing on only \textbf{female} population claim that `\textit{COVID-19 vaccine is safe during pregnancy and breastfeeding}'. `\textbf{EncourageVaccination}' themed ads only target \textbf{female} population with mentioned entities `\textbf{\textit{family}}', `\textbf{\textit{kids}}', `\textbf{\textit{community}}', `\textbf{\textit{friends}}' and emphasize that `\textit{vaccination is the best way to protect kids, family, friends, and community}'. On the other hand, advertisements having `\textbf{VaccineMisinformation}' target \textbf{female} population by claiming `\textit{COVID-19 vaccine does not contain aborted fetal cells, microchips, mercury, formaldehyde and mRNA can't alter DNA}'.
\begin{table}[t]
\begin{center}
 \scalebox{0.95}{\begin{tabular}{>{\arraybackslash}m{1.1cm}|>{\arraybackslash}m{7cm}}
 \toprule
 \textsc{\textbf{Pro-energy}} & Corporation: \textbf{Exxon Mobil}, Industry Association: \textbf{American Petroleum Institute}, Advocacy Group: \textbf{Consumer Energy Alliance}. \\
 \hline
 \textsc{\textbf{Clean-energy}} & Corporation: \textbf{Apex clean energy}, Industry Association: \textbf{American Wind Energy Association}, Advocacy Group: \textbf{Climate Power Education Fund}.\\
 \bottomrule
\end{tabular}}
\caption{List of funding entities from Climate dataset.}
\label{tab:fe}
\end{center}
\end{table}
\subsection{Geographic Variations in Climate Themes}
To grasp the ways in which fossil fuel industries and their supporting groups shape public opinion, \citet{islam2023analysis} categorize \textit{pro-energy} funding entities into three distinct types: \textbf{Corporation}, \textbf{Industry Association}, and \textbf{Advocacy Group}. Similarly, we categorize the \textit{clean-energy} funding entities into those categories. In this paper, we choose only one funding entity from each category resulting $6$ funding entities, including $3$ pro-energy and $3$ clean-energy funding entities. Table \ref{tab:fe} shows the list of funding entities included in our analysis. 

We divide the USA into $5$ regions, encompassing all $50$ states: West, Southwest, Midwest, Southeast, and Northeast to analyze the geographic targeting of the sponsored ads (Fig. \ref{fig:geo_fe}). \textbf{Exxon mobile} targets \textbf{\textit{Northeast}} region sponsoring ads focused on `\textit{energy affordability and reliability}' (`\textbf{Pragmatism}' theme). For the rest of the regions, it sponsor ads focusing on `\textit{how energy can play in supporting jobs, wages and community growth}' (`\textbf{Economy\_pro}' theme).
On the other hand, the pro-energy advocacy group \textbf{Consumer Energy Alliance} focuses on the `\textbf{Economy\_pro}' theme when targeting the \textbf{\textit{Midwest}}, but shifts to the `\textbf{ClimateSolution}' theme for the \textbf{\textit{Southeast}} region (Fig. \ref{fig:cons}).

From Fig. \ref{fig:apx}, we notice that \textbf{Apex clean energy} targets the \textbf{\textit{West}} region focusing on `\textbf{EcofriendlyConsumerChoices}' and `\textbf{PromoteSustainableTransportation}' themes, while for other regions (except the Southwest) it emphasizes on `\textbf{CustomerBasedAltEnergy}' theme. However, \textbf{American Wind Energy Association} uses `\textbf{PoliticalAffiliation}' theme to target the \textbf{\textit{Southwest}} and \textbf{\textit{Midwest}} regions (Fig. \ref{fig:wind}).
\section {Theme Shifts Triggered by Key Events}
To investigate how talking points change in response to real-world events, we pick $2$ defining events ($1$ event per case study) from our case studies. The usage of themes in response to these events is shown in Fig. \ref{fig:event}. The events are as follows:
\begin{itemize}
 \item \textbf{Gas Price Increase:} On May $12$, $2021$, US national average retail gas prices rose above $\$3$ a gallon for the first time since $2014$\footnote{\url{https://www.bloomberg.com/news/articles/2021-05-12/u-s-gasoline-prices-rise-above-3-gallon-first-time-since-2014}}.
 
  \item \textbf{Federal Vaccine Mandate:} On September $09$, $2021$, President Biden announced an executive order on COVID-19 vaccine mandates for federal workers, large employers, and health care staff\footnote{\url{https://www.whitehouse.gov/briefing-room/presidential-actions/2021/09/09/executive-order-on-requiring-coronavirus-disease-2019-vaccination-for-federal-employees/}}.
\end{itemize}
Fig. \ref{fig:gp} displays the top-four themes in sponsored advertisements three days before and after the significant event labeled as the \textbf{Gas Price Increase} on May $12$, $2021$. Following this event, there is a noticeable increase in the frequency of themes such as `\textbf{AltEnergy}' and `\textbf{ClimateSolution}'. Additionally, themes like `\textbf{BidenGasPriceIncrease}' and `\textbf{AgainstCorporateInterests}' become prominently visible in the aftermath.

Fig. \ref{fig:vax} displays the thematic shift in sponsored advertisements before and after President Biden announced the \textbf{Federal vaccine mandate} on September $09$, $2021$. Post-announcement, themes such as `\textbf{GovDistrust}' and `\textbf{VaccineMisinformation}' indicate increasing concerns about the trustworthiness of the COVID-19 vaccine. Additionally, there is a notable increase in themes directly related to the \textbf{vaccine mandate}.
\vspace{-5pt}
\section{Conclusion}
We present a \textbf{machine-in-the-loop} framework for discovering latent themes from social media messaging, which can help speed up the development of domain-specific labels. We show that themes generated from our approach can cover a larger portion of ads, and those themes are coherent through quantitative measures and human evaluation. Additionally, we demonstrate that our resulting set of themes is cleaner and more explainable than themes obtained with topic modeling approaches. Furthermore, our analysis demonstrates how these themes are specifically tailored for demographic targeting on social media and how the thematic focus of messaging shifts in response to real-world events.
\vspace{-5pt}
\section{Limitations}
Though we show the effectiveness of our framework focusing on two case studies, i.e., climate and COVID-19 vaccine campaigns, it is a domain-agnostic framework. However, the main idea of characterizing themes using LLMs and associating the ads to their closest themes is applicable to other cases with no changes.

Our approach relies on GPT-4 to generate a multi-document summary, check cluster coherency, and map ads to themes. We chose GPT-4 instead of the
open-source counterparts due to computational resource constraints. 


LLMs are pre-trained on a huge amount of
human-generated text. As a result, they may
inherently contain many human biases \cite{blodgett2020language,brown2020language}. We did not
consider any bias in our task.

A limitation of this study is the inherent subjectivity in interpreting demographic targeting, as the differentiation of themes based on demographics (specifically age and gender in this paper) is subjective, which can affect the utility of research findings.
\vspace{-5pt}
\section{Ethics Statement}
To the best of our knowledge, we did not violate any ethical code while conducting the research work described in this paper. We report the technical details for the reproducibility of results. 

In this paper, we did not introduce any new dataset; instead, we experimented using existing datasets that are adequately cited.

The author's personal views are not represented in any qualitative results we report, as it is solely an outcome derived from a machine learning model.
\vspace{-5pt}
\section{Broader Impact}
Our approach marks a significant advancement in the thematic analysis of social media messaging, which might provide valuable insights into the strategic use of messaging for targeted communication and campaign efficacy.
We leave this exploration for future work.
\vspace{-5pt}
\bibliography{aaai24}

\begin{thebibliography}{58}
\providecommand{\natexlab}[1]{#1}

\bibitem[{Achiam et~al.(2023)}]{achiam2023gpt}
Achiam, J.; et~al. 2023.
\newblock Gpt-4 technical report.
\newblock \emph{arXiv preprint arXiv:2303.08774}.

\bibitem[{Andreou et~al.(2019)}]{andreou2019measuring}
Andreou, A.; et~al. 2019.
\newblock Measuring the Facebook advertising ecosystem.
\newblock In \emph{NDSS}.

\bibitem[{Barbu(2014)}]{barbu2014advertising}
Barbu, O. 2014.
\newblock Advertising, microtargeting and social media.
\newblock \emph{Procedia-Social and Behavioral Sciences}.

\bibitem[{Blei, Ng, and Jordan(2003)}]{blei2003latent}
Blei, D.~M.; Ng, A.~Y.; and Jordan, M.~I. 2003.
\newblock Latent dirichlet allocation.
\newblock \emph{JMLR}.

\bibitem[{Blodgett et~al.(2020)Blodgett, Barocas, Daum{\'e}~III, and Wallach}]{blodgett2020language}
Blodgett, S.~L.; Barocas, S.; Daum{\'e}~III, H.; and Wallach, H. 2020.
\newblock Language (Technology) is Power: A Critical Survey of “Bias” in NLP.
\newblock In \emph{ACL}.

\bibitem[{Braun and Clarke(2012)}]{braun2012thematic}
Braun, V.; and Clarke, V. 2012.
\newblock \emph{Thematic analysis.}
\newblock APA.

\bibitem[{Braun and et. al(2006)}]{braun2006using}
Braun, V.; and et. al. 2006.
\newblock Using thematic analysis in psychology.
\newblock \emph{Qual. Res. Psychol.}

\bibitem[{Brown et~al.(2020)}]{brown2020language}
Brown, T.; et~al. 2020.
\newblock Language models are few-shot learners.
\newblock \emph{NeurIPS}.

\bibitem[{Capozzi et~al.(2020)}]{capozzi2020facebook}
Capozzi, A.; et~al. 2020.
\newblock Facebook Ads: Politics of Migration in Italy.
\newblock In \emph{ICSI}.

\bibitem[{Capozzi et~al.(2021)}]{capozzi2021clandestino}
Capozzi, A.; et~al. 2021.
\newblock Clandestino or Rifugiato? Anti-immigration Facebook Ad Targeting in Italy.
\newblock In \emph{CHI}.

\bibitem[{Chang et~al.(2009)}]{chang2009reading}
Chang, J.; et~al. 2009.
\newblock Reading tea leaves: How humans interpret topic models.
\newblock \emph{NeurIPS}.

\bibitem[{Chiang and Lee(2023)}]{chiang2023can}
Chiang, C.-H.; and Lee, H.-y. 2023.
\newblock Can Large Language Models Be an Alternative to Human Evaluations?
\newblock In \emph{ACL}.

\bibitem[{Chowdhery et~al.(2023)}]{chowdhery2023palm}
Chowdhery, A.; et~al. 2023.
\newblock Palm: Scaling language modeling with pathways.
\newblock \emph{JMLR}.

\bibitem[{Dai et~al.(2023)}]{dai2023llm}
Dai, S.-C.; et~al. 2023.
\newblock LLM-in-the-loop: Leveraging Large Language Model for Thematic Analysis.
\newblock \emph{arXiv preprint arXiv:2310.15100}.

\bibitem[{De~Paoli(2023)}]{de2023can}
De~Paoli, S. 2023.
\newblock Can Large Language Models emulate an inductive Thematic Analysis of semi-structured interviews? An exploration and provocation on the limits of the approach and the model.
\newblock \emph{arXiv preprint arXiv:2305.13014}.

\bibitem[{Fan et~al.(2014)}]{fan2014challenges}
Fan, J.; et~al. 2014.
\newblock Challenges of big data analysis.
\newblock \emph{NSR}.

\bibitem[{{FORCE11}(2020)}]{fair}
{FORCE11}. 2020.
\newblock The FAIR Data principles.
\newblock \url{https://force11.org/info/the-fair-data-principles/}.

\bibitem[{Gao and Guo(2024)}]{gao2023collabcoder}
Gao, J.; and Guo, o. 2024.
\newblock CollabCoder: A GPT-Powered Workflow for Collaborative Qualitative Analysis.
\newblock \emph{CHI}.

\bibitem[{Gebru et~al.(2021)Gebru, Morgenstern, Vecchione, Vaughan, Wallach, Iii, and Crawford}]{gebru2021datasheets}
Gebru, T.; Morgenstern, J.; Vecchione, B.; Vaughan, J.~W.; Wallach, H.; Iii, H.~D.; and Crawford, K. 2021.
\newblock Datasheets for datasets.
\newblock \emph{Communications of the ACM}, 64(12): 86--92.

\bibitem[{Gilardi, Alizadeh, and Kubli(2023)}]{gilardi2023chatgpt}
Gilardi, F.; Alizadeh, M.; and Kubli, M. 2023.
\newblock ChatGPT outperforms crowd workers for text-annotation tasks.
\newblock \emph{Proceedings of the National Academy of Sciences}, 120(30): e2305016120.

\bibitem[{Grootendorst(2022)}]{grootendorst2022bertopic}
Grootendorst, M. 2022.
\newblock BERTopic: Neural topic modeling with a class-based TF-IDF procedure.
\newblock \emph{arXiv preprint arXiv:2203.05794}.

\bibitem[{Haidt and Graham(2007)}]{haidt2007morality}
Haidt, J.; and Graham, J. 2007.
\newblock When morality opposes justice: Conservatives have moral intuitions that liberals may not recognize.
\newblock \emph{SJR}.

\bibitem[{Haidt and Joseph(2004)}]{haidt2004intuitive}
Haidt, J.; and Joseph, C. 2004.
\newblock Intuitive ethics: How innately prepared intuitions generate culturally variable virtues.
\newblock \emph{Daedalus}, 133(4): 55--66.

\bibitem[{Hersh(2015)}]{hersh2015}
Hersh, E.~D. 2015.
\newblock \emph{Hacking the electorate: How campaigns perceive voters}.
\newblock Cambridge University Press.

\bibitem[{Hoyle et~al.(2021)}]{hoyle2021automated}
Hoyle, A.; et~al. 2021.
\newblock Is automated topic model evaluation broken? the incoherence of coherence.
\newblock \emph{NeurIPS}.

\bibitem[{Hu et~al.(2011)}]{hu2014interactive}
Hu, Y.; et~al. 2011.
\newblock Interactive topic modeling.
\newblock \emph{ACL}.

\bibitem[{Islam and et. al(2023)}]{islam2023weakly}
Islam, T.; and et. al. 2023.
\newblock Weakly Supervised Learning for Analyzing Political Campaigns on Facebook.
\newblock In \emph{ICWSM}.

\bibitem[{Islam and Goldwasser(2022{\natexlab{a}})}]{islam2022understanding}
Islam, T.; and Goldwasser, D. 2022{\natexlab{a}}.
\newblock Understanding COVID-19 Vaccine Campaign on Facebook using Minimal Supervision.
\newblock In \emph{2022 IEEE International Conference on Big Data (Big Data)}, 585--595. IEEE.

\bibitem[{Islam and Goldwasser(2022{\natexlab{b}})}]{islam2022covidfbAd}
Islam, T.; and Goldwasser, D. 2022{\natexlab{b}}.
\newblock Understanding COVID-19 Vaccine Campaign on Facebook using Minimal Supervision.
\newblock In \emph{IEEE Big Data}.

\bibitem[{Islam et~al.(2023)}]{islam2023analysis}
Islam, T.; et~al. 2023.
\newblock Analysis of Climate Campaigns on Social Media Using Bayesian Model Averaging.
\newblock In \emph{AIES}.

\bibitem[{Jagadish et~al.(2014)}]{jagadish2014big}
Jagadish, H.~V.; et~al. 2014.
\newblock Big data and its technical challenges.
\newblock \emph{CACM}.

\bibitem[{Jin and Han(2010)}]{Jin2010}
Jin, X.; and Han, J. 2010.
\newblock \emph{K-Means Clustering}.

\bibitem[{Joachims(1998)}]{joachims1998text}
Joachims, T. 1998.
\newblock Text categorization with support vector machines: Learning with many relevant features.
\newblock In \emph{ECML}.

\bibitem[{Kojima et~al.(2022)}]{kojima2022large}
Kojima, T.; et~al. 2022.
\newblock Large language models are zero-shot reasoners.
\newblock \emph{NeurIPS}.

\bibitem[{Lau et~al.(2014)}]{lau2014machine}
Lau, J.~H.; et~al. 2014.
\newblock Machine reading tea leaves: Automatically evaluating topic coherence and topic model quality.
\newblock In \emph{EACL}.

\bibitem[{Le~Scao et~al.(2022)}]{le2022bloom}
Le~Scao, T.; et~al. 2022.
\newblock Bloom: A 176b-parameter open-access multilingual language model.

\bibitem[{LeCun, Bengio, and Hinton(2015)}]{lecun2015deep}
LeCun, Y.; Bengio, Y.; and Hinton, G. 2015.
\newblock Deep learning.
\newblock \emph{nature}.

\bibitem[{Lee and Seung(1999)}]{lee1999learning}
Lee, D.~D.; and Seung, H.~S. 1999.
\newblock Learning the parts of objects by non-negative matrix factorization.
\newblock \emph{Nature}, 401(6755): 788--791.

\bibitem[{Lund et~al.(2017)}]{lund2017tandem}
Lund, J.; et~al. 2017.
\newblock Tandem anchoring: A multiword anchor approach for interactive topic modeling.
\newblock In \emph{ACL}.

\bibitem[{McCallum et~al.(1998)}]{mccallum1998comparison}
McCallum, A.; et~al. 1998.
\newblock A comparison of event models for naive bayes text classification.
\newblock In \emph{Learning for text categorization}.

\bibitem[{Mejova and Kalimeri(2020)}]{mejova2020covid}
Mejova, Y.; and Kalimeri, K. 2020.
\newblock COVID-19 on Facebook ads: competing agendas around a public health crisis.
\newblock In \emph{COMPASS}.

\bibitem[{Mimno et~al.(2011)}]{mimno2011optimizing}
Mimno, D.; et~al. 2011.
\newblock Optimizing semantic coherence in topic models.
\newblock In \emph{EMNL}.

\bibitem[{Pacheco, Islam et~al.(2022{\natexlab{a}})}]{pacheco2022holistic}
Pacheco, M.~L.; Islam, T.; et~al. 2022{\natexlab{a}}.
\newblock A Holistic Framework for Analyzing the COVID-19 Vaccine Debate.
\newblock In \emph{NAACL}.

\bibitem[{Pacheco, Islam et~al.(2022{\natexlab{b}})}]{pacheco2022interactively}
Pacheco, M.~L.; Islam, T.; et~al. 2022{\natexlab{b}}.
\newblock Interactively uncovering latent arguments in social media platforms: A case study on the covid-19 vaccine debate.
\newblock In \emph{DASH}.

\bibitem[{Pacheco, Islam et~al.(2023)}]{pacheco2023interactive}
Pacheco, M.~L.; Islam, T.; et~al. 2023.
\newblock Interactive Concept Learning for Uncovering Latent Themes in Large Text Collections.
\newblock In \emph{ACL Findings}.

\bibitem[{Rehurek and Sojka(2011)}]{rehurek2011gensim}
Rehurek, R.; and Sojka, P. 2011.
\newblock Gensim--python framework for vector space modelling.

\bibitem[{Reimers and Gurevych(2019)}]{reimers2019sentence}
Reimers, N.; and Gurevych, I. 2019.
\newblock Sentence-{BERT}: Sentence Embeddings using {S}iamese {BERT}-Networks.
\newblock In \emph{EMNLP-IJCNLP}.

\bibitem[{Roberts et~al.(2019)}]{roberts2019attempting}
Roberts, K.; et~al. 2019.
\newblock Attempting rigour and replicability in thematic analysis of qualitative research data; a case study of codebook development.
\newblock \emph{BMC Med. Res. Methodol.}

\bibitem[{R{\"o}der et~al.(2015)}]{roder2015exploring}
R{\"o}der, M.; et~al. 2015.
\newblock Exploring the space of topic coherence measures.
\newblock In \emph{WSDM}.

\bibitem[{Serrano et~al.(2020)}]{serrano2020political}
Serrano, J. C.~M.; et~al. 2020.
\newblock The Political Dashboard: A Tool for Online Political Transparency.
\newblock In \emph{ICWSM}.

\bibitem[{Silva and Benevenuto(2021)}]{silva2021covid}
Silva, M.; and Benevenuto, F. 2021.
\newblock COVID-19 ads as political weapon.
\newblock In \emph{SAC}.

\bibitem[{Silva et~al.(2020)}]{silva2020facebook}
Silva, M.; et~al. 2020.
\newblock Facebook Ads Monitor: An Independent Auditing System for Political Ads on Facebook.
\newblock In \emph{WWW}.

\bibitem[{Sivarajah et~al.(2017)}]{sivarajah2017critical}
Sivarajah, U.; et~al. 2017.
\newblock Critical analysis of Big Data challenges and analytical methods.
\newblock \emph{Journal of business research}.

\bibitem[{Tuckett(2005)}]{tuckett2005applying}
Tuckett, A.~G. 2005.
\newblock Applying thematic analysis theory to practice: A researcher’s experience.
\newblock \emph{Contemporary nurse}.

\bibitem[{Vaismoradi et~al.(2013)}]{vaismoradi2013content}
Vaismoradi, M.; et~al. 2013.
\newblock Content analysis and thematic analysis: Implications for conducting a qualitative descriptive study.
\newblock \emph{Nursing \& health sciences}.

\bibitem[{Vaismoradi et~al.(2016)}]{vaismoradi2016theme}
Vaismoradi, M.; et~al. 2016.
\newblock Theme development in qualitative content analysis and thematic analysis.

\bibitem[{Xiao et~al.(2023)}]{xiao2023supporting}
Xiao, Z.; et~al. 2023.
\newblock Supporting Qualitative Analysis with Large Language Models: Combining Codebook with GPT-3 for Deductive Coding.
\newblock In \emph{IUI}.

\bibitem[{Ziems et~al.(2024)}]{ziems2024can}
Ziems, C.; et~al. 2024.
\newblock Can large language models transform computational social science?
\newblock \emph{Computational Linguistics}.

\end{thebibliography}

\subsection{Paper Checklist to be included in your paper}

\begin{enumerate}
\item For most authors...
\begin{enumerate}
    \item  Would answering this research question advance science without violating social contracts, such as violating privacy norms, perpetuating unfair profiling, exacerbating the socio-economic divide, or implying disrespect to societies or cultures?
    \answerTODO{Answer} Yes
  \item Do your main claims in the abstract and introduction accurately reflect the paper's contributions and scope?
    \answerTODO{Answer} Yes
   \item Do you clarify how the proposed methodological approach is appropriate for the claims made? 
    \answerTODO{Answer} Yes
   \item Do you clarify what are possible artifacts in the data used, given population-specific distributions?
    \answerTODO{Answer} NA
  \item Did you describe the limitations of your work?
    \answerTODO{Answer} YES
  \item Did you discuss any potential negative societal impacts of your work?
    \answerTODO{Answer} NA
      \item Did you discuss any potential misuse of your work? 
    \answerTODO{Answer} NA
    \item Did you describe steps taken to prevent or mitigate potential negative outcomes of the research, such as data and model documentation, data anonymization, responsible release, access control, and the reproducibility of findings? 
    \answerTODO{Answer} Yes
  \item Have you read the ethics review guidelines and ensured that your paper conforms to them?
    \answerTODO{Answer} Yes
\end{enumerate}

\item Additionally, if your study involves hypotheses testing...
\begin{enumerate}
  \item Did you clearly state the assumptions underlying all theoretical results?
    \answerTODO{Answer} NA
  \item Have you provided justifications for all theoretical results?
    \answerTODO{Answer}
  \item Did you discuss competing hypotheses or theories that might challenge or complement your theoretical results?
    \answerTODO{Answer} NA
  \item Have you considered alternative mechanisms or explanations that might account for the same outcomes observed in your study?
    \answerTODO{Answer} NA
  \item Did you address potential biases or limitations in your theoretical framework?
    \answerTODO{Answer} NA
  \item Have you related your theoretical results to the existing literature in social science?
    \answerTODO{Answer} NA
  \item Did you discuss the implications of your theoretical results for policy, practice, or further research in the social science domain?
    \answerTODO{Answer} NA
\end{enumerate}

\item Additionally, if you are including theoretical proofs...
\begin{enumerate}
  \item Did you state the full set of assumptions of all theoretical results?
    \answerTODO{Answer} NA
	\item Did you include complete proofs of all theoretical results?
    \answerTODO{Answer} NA
\end{enumerate}

\item Additionally, if you ran machine learning experiments...
\begin{enumerate}
  \item Did you include the code, data, and instructions needed to reproduce the main experimental results (either in the supplemental material or as a URL)?
    \answerTODO{Answer} Yes
  \item Did you specify all the training details (e.g., data splits, hyperparameters, how they were chosen)?
    \answerTODO{Answer} Yes
     \item Did you report error bars (e.g., with respect to the random seed after running experiments multiple times)?
    \answerTODO{Answer} NA
	\item Did you include the total amount of compute and the type of resources used (e.g., type of GPUs, internal cluster, or cloud provider)? 
    \answerTODO{Answer} Yes
     \item Do you justify how the proposed evaluation is sufficient and appropriate to the claims made? 
    \answerTODO{Answer} Yes
     \item Do you discuss what is ``the cost`` of misclassification and fault (in)tolerance?
    \answerTODO{Answer} Yes
  
\end{enumerate}

\item Additionally, if you are using existing assets (e.g., code, data, models) or curating/releasing new assets, \textbf{without compromising anonymity}...
\begin{enumerate}
  \item If your work uses existing assets, did you cite the creators?
    \answerTODO{Answer} Yes
  \item Did you mention the license of the assets?
    \answerTODO{Answer} NA
  \item Did you include any new assets in the supplemental material or as a URL?
    \answerTODO{Answer} No, because I used existing dataset.
  \item Did you discuss whether and how consent was obtained from people whose data you're using/curating?
    \answerTODO{Answer} Yes
  \item Did you discuss whether the data you are using/curating contains personally identifiable information or offensive content?
    \answerTODO{Answer} Yes
\item If you are curating or releasing new datasets, did you discuss how you intend to make your datasets FAIR (see \citet{fair})?
\answerTODO{Answer} NA
\item If you are curating or releasing new datasets, did you create a Datasheet for the Dataset (see \citet{gebru2021datasheets})? 
\answerTODO{Answer} NA
\end{enumerate}

\item Additionally, if you used crowdsourcing or conducted research with human subjects, \textbf{without compromising anonymity}...
\begin{enumerate}
  \item Did you include the full text of instructions given to participants and screenshots?
    \answerTODO{Answer} NA
  \item Did you describe any potential participant risks, with mentions of Institutional Review Board (IRB) approvals? 
    \answerTODO{Answer} NA
  \item Did you include the estimated hourly wage paid to participants and the total amount spent on participant compensation?
    \answerTODO{Answer} NA
   \item Did you discuss how data is stored, shared, and deidentified?
   \answerTODO{Answer} NA
\end{enumerate}

\end{enumerate}

\appendix
\section{Appendix}
\label{sec:appendix}
\subsection{Climate Campaigns Cluster Summaries}
\label{sec:cl_sum_climate}
Summaries of the final $25$ themes (clusters) from the climate case study are shown below:

\noindent \textbf{1. Summary\_Economy\_pro: }
The provided texts collectively emphasize the importance of the oil and natural gas industry in supporting local jobs, economic growth, and community development. They express concerns about proposed bans on natural gas and oil, warning of potential job losses and negative economic impacts. The texts call for support of the oil and gas industry, highlighting its contributions to local economies and communities while advocating for a sustainable future.

\noindent \textbf{2. Summary\_ClimateSolution: }
The provided texts collectively advocate for the role of natural gas in advancing a cleaner and more sustainable energy future. They highlight natural gas as an essential tool for reducing carbon emissions, providing affordable and reliable energy, and complementing renewable energy sources like wind and solar. The texts emphasize the importance of using natural gas to address the dual challenge of meeting energy needs and protecting the environment while working towards ambitious carbon reduction goals, such as reaching net-zero emissions by 2045. Texts focus on greenwashing aspect and the promotion of a controversial energy source under the guise of environmental benefit. The text covers the complex and sometimes contradictory narratives around certain fossil fuels in the context of climate and energy policy.

\noindent \textbf{3. Summary\_Pragmatism:}
The provided texts collectively stress the vital role of reliable and affordable natural gas and oil in enabling modern life. They highlight that in 2020, natural gas and oil accounted for nearly $70\%$ of America's energy supply, and projections suggest they will continue to meet approximately $70\%$ of the nation's energy needs by 2050, as indicated by the U.S. Energy Information Administration. The texts call attention to the significance of American oil and gas production in maintaining affordable energy and invite individuals to share their priorities in this regard.

\noindent \textbf{4. Summary\_Patriotism:}
The provided texts collectively emphasize the importance of U.S. energy leadership and innovation, highlighting the country's position as a top global energy producer while actively working to reduce greenhouse gas emissions. They express concerns about potential cutbacks in local oil and gas production, citing potential consequences such as increased reliance on foreign oil imports and higher gas prices for Californians. The texts argue against shutting off American oil and natural gas production, pointing out that such actions could strengthen other nations while negatively impacting U.S. jobs, economic security, and national security. Additionally, they underscore the broader impacts of U.S. oil and natural gas production on daily life, including its role in fertilizer production and its significance in ensuring food supply and availability. The texts call for action to protect America's energy future and support natural gas and consumer choice through petitions, and they highlight the U.S.'s achievement as the world's No. 1 exporter of liquefied natural gas, particularly in response to energy demands in Europe.

\noindent \textbf{5. Summary\_AgainstClimatePolicy:}
The provided texts collectively convey a sense of urgency and concern regarding the oil and gas industry, emphasizing the need for support and resistance against what they perceive as threats to the industry's survival. They criticize progressive and Democratic energy policies, particularly the Green New Deal, as potentially devastating to American workers and the economy. The texts also criticize the Biden administration's energy policies, linking them to high gas prices and a reliance on emergency reserves. They call for action to prioritize and protect the American energy industry, with an emphasis on opposing what they view as radical policies and mandates. Additionally, the texts highlight the role of the oil and gas industry in addressing climate challenges through innovation and expertise while expressing fears of job loss and economic consequences if certain policies are implemented.

\noindent \textbf{6. Summary\_Economy\_clean:}
The provided texts collectively advocate for a clean energy future in the United States and emphasize the potential for job creation and economic benefits associated with clean energy initiatives. They highlight the growth of clean energy sectors and argue that investments in renewable energy are generating good-paying jobs across various occupations and regions. The texts contrast clean energy with fossil fuels, suggesting that clean energy is not only environmentally responsible but also economically advantageous. They call on Congress to support clean energy policies and investments as a means to strengthen the American economy, create jobs, lower energy costs, and mitigate climate change, ultimately promoting a clean energy-focused approach to economic growth.

\noindent \textbf{7. Summary\_HumanHealth:}
The provided texts collectively emphasize the significant and urgent health impacts of climate change, highlighting the range of dangers it poses, including cardiovascular and respiratory issues, food security, the spread of diseases, and more. They stress the need for awareness and action to address these health challenges linked to climate change, with mentions of governmental initiatives and organizations dedicated to addressing climate-related health concerns. The texts underscore that climate change is already harming people and communities, affecting vulnerable populations, and call for support and action to mitigate these adverse health effects.

\noindent \textbf{8. Summary\_FutureGeneration:}
The provided texts collectively emphasize the grave threat of climate change to children's health and well-being. They cite alarming statistics, including data from the World Health Organization, that underscore the vulnerability of young children to climate-related health issues. The texts stress the urgent need to protect children from the impacts of climate change and highlight the various health risks they face, including premature births, birth defects, and increased emergency room visits. They call for action to address climate change to safeguard the future of children and grandchildren, emphasizing that inaction comes at a high cost to children's health and well-being. The texts also acknowledge the anxiety and concerns that climate change is causing among young people and suggest ways to support and address their anxieties.

\noindent \textbf{9. Summary\_Environmental:}
The provided texts collectively address the urgency and breadth of the climate crisis, emphasizing its far-reaching impacts on various aspects of society and the environment. They highlight alarming trends such as the rapid warming of oceans, the increasing frequency of extreme weather events, and the dire consequences of climate change on ecosystems, economies, and public health. The texts call for collective action and emphasize the role of individuals in combating climate change, urging the sharing of ideas and solutions. They also stress the need to address the root causes, including greenhouse gas emissions from fossil fuels, and promote carbon capture as a key strategy for decarbonizing the global energy system. Overall, the texts underscore the severity of the climate emergency and the importance of taking immediate action to mitigate its effects and protect vulnerable communities and ecosystems.

\noindent \textbf{10. Summary\_Animals:}
The provided texts collectively highlight the severe impact of climate change on wildlife, fisheries, birds and ecosystems. They mention rising temperatures, increased droughts, fires, intense storms, and floods that threaten various species and their habitats. Studies reveal that climate change could lead to the extinction of one-third of species in the next 50 years, with particular vulnerabilities in South America. The texts emphasize the urgent need for action to protect wildlife from the consequences of climate change, emphasizing the role of renewable energy development that must consider the needs of wildlife. They call for an understanding of the interconnected crises of climate change and biodiversity loss, urging collective efforts to address both challenges.

\noindent \textbf{11. Summary\_AltEnergy:  }
The provided texts collectively advocate for the adoption of renewable energy sources as a cleaner, cheaper, and more sustainable alternative to fossil fuels. They emphasize the benefits of transitioning to renewables, including reducing carbon footprints, enhancing grid reliability, and mitigating the impact of rising gas prices. The texts highlight the rapid growth of renewable energy in the United States, with clean power sources like wind, solar, and hydrogen accounting for a significant share of electricity generation. They stress the urgency of making this transition to combat climate change, citing the irreversible impacts looming by 2030. Additionally, they mention various projects and initiatives aimed at promoting clean energy and energy independence, such as repurposing coal plants into renewable energy hubs and proposed solar farms. Finally, they encourage energy-conscious practices like using electricity during off-peak hours to harness clean energy effectively.

\noindent \textbf{12. Summary\_SupportClimatePolicy:  }
The provided texts collectively advocate for the passage of the Build Back Better Act, emphasizing its significance in addressing climate change and promoting a clean energy future. They highlight the key provisions of the bill, including cutting pollution by $50\%$ by 2030, creating clean energy jobs, and lowering utility bills. The texts stress the urgency of taking action to combat climate change and emphasize that it's a historic opportunity to invest in meaningful climate action and transition to a carbon-zero grid. Additionally, some texts criticize the fossil fuel industry for its role in climate change and call for holding polluters accountable. Overall, the focus is on the importance of the Build Back Better Act in advancing climate goals and environmental justice.

\noindent \textbf{13. Summary\_PoliticalAffiliation:}
The provided texts collectively highlight concerns about the influence of the oil and gas industry on climate change policies. They emphasize that despite claims of caring about climate change, these industries spend significant amounts on lobbying to block regulations and protect their subsidies. The texts mention the substantial sums spent on climate lobbying by major oil and gas companies and argue that these actions hinder meaningful climate action. They also discuss political contributions to lawmakers and emphasize the need to address climate change without being influenced by industry lobbying. Additionally, some texts reference specific instances where lobbyists have allegedly undermined climate legislation or influenced infrastructure bills to prioritize industry interests over environmental concerns. Overall, the focus is on revealing the industry's efforts to shape climate policies to their advantage.

\noindent \textbf{14. Summary\_BidenGasPriceIncrease: } 
The provided texts collectively express strong disapproval and criticism of President Joe Biden's administration, focusing on various issues such as rising gas prices, inflation, the southern border crisis, and energy policies. Some texts call for Joe Biden's resignation and invite readers to participate in approval polls, while others highlight specific concerns, like the increase in gas prices and the cancellation of certain pipelines, portraying these actions as detrimental to the American economy. Overall, these texts present a negative view of Joe Biden's leadership and policy decisions.

\noindent \textbf{15. Summary\_AgainstCorporateInterests:}
The provided texts largely revolve around progressive candidates running for Congress and their calls for grassroots support. These candidates emphasize their commitment to economic justice, racial equity, climate action, and progressive policies such as a Green New Deal, Medicare for All, and universal housing. They criticize their opponents for receiving significant contributions from corporations and lobbyists and highlight the need for people-powered campaigns. Some texts also mention the importance of supporting incumbent progressive Congress members, such as the Squad, to pass transformative legislation like the Build Back Better Agenda. Overall, these texts aim to rally support for progressive candidates and causes while advocating for a shift away from corporate influence in politics.  

\noindent \textbf{16. Summary\_GasTax:}
The provided texts revolve around opposing proposed gas tax increases in various states. They emphasize concerns about the financial burden on families, rising gas prices, and the potential for gasoline shortages. The texts encourage people to sign petitions or participate in rallies to voice their opposition to these tax hikes. Additionally, some texts highlight the current tax rates and the impact of previous tax increases, framing the issue as a matter of transparency and affordability for consumers. Overall, the focus is on mobilizing public sentiment against gas tax increases in different states.

\noindent \textbf{17. Summary\_Deforestation:}
The texts primarily focus on environmental conservation and efforts to combat climate change. They highlight the importance of preserving forests like the Tongass National Forest in Alaska, emphasizing their role in storing carbon, providing habitat, and mitigating climate change effects. They urge readers to take action, such as signing petitions or participating in tree-planting initiatives. Some texts discuss the potential benefits of "super trees" that are more effective at carbon storage and stress the importance of preserving mature trees in the face of climate change. Overall, the texts call for collective action to protect and restore natural environments and address climate-related challenges by planting more trees and preserving forests.

\noindent \textbf{18. Summary\_Carbon:}
These texts collectively emphasize the importance of reducing carbon emissions and mitigating one's carbon footprint. They introduce various strategies and initiatives for individuals and communities to lower their carbon impact, such as carbon pricing, using energy-efficient lights, enrolling in carbon reduction programs, calculating personal carbon footprints, offsetting tailpipe emissions, and supporting sustainability efforts in industries like cement and concrete. The overarching theme is the need for environmental awareness and action to combat climate change and promote a more sustainable future.

\noindent \textbf{19. Summary\_CustomerBasedAltEnergy:}
These texts collectively highlight various aspects of renewable and sustainable energy sources. They address common myths and misconceptions about solar energy, emphasizing its versatility and suitability for different types of buildings and applications. They promote community solar programs in Illinois, offer information and resources for learning about solar power, and invite individuals to explore solar technology and installations. Additionally, the texts touch upon other forms of renewable energy, such as wind and nuclear power, showcasing their potential to provide reliable, clean energy and create jobs in various states. The overarching theme is the promotion and education about renewable energy sources and their benefits.

\noindent \textbf{20. Summary\_FoodSecurity:}
The provided texts collectively address the complexities of food security, emphasizing the global nature of the challenge and the need for cross-sector collaboration to meet the increasing food demand caused by a growing global population and climate change. They also shed light on issues of food justice and sovereignty, particularly in marginalized communities, where access to healthy foods is limited. Additionally, the texts discuss the concept of food elitism, questioning marketing practices and their impact on food equity, arguing for the importance of affordable and sustainable food systems. Furthermore, the issue of food deserts is explored, where residents face challenges related to the availability, accessibility, and affordability of nutritious food, with a recognition of the intricate factors contributing to food insecurity. These texts collectively stress the urgency of addressing food-related challenges from multiple perspectives, encompassing policy, social justice, and sustainable agriculture.

\noindent \textbf{21. Summary\_EnergyAffordabilityandSustainabilityLegislation:}
These texts all convey similar messages, focusing on various representatives, including Rep. John Autry, Rep. Kelly Alexander, Rep. Kyle Hall, Rep. John Sauls, and Rep. John A. Torbett, who are collectively working to reduce power bills for families and businesses in North Carolina. They sponsor legislation aimed at exploring changes that could potentially result in annual energy bill reductions exceeding $\$600$ million. The texts encourage readers to learn more about the Alliance for Energy Security's collaboration with policymakers in an effort to make energy in North Carolina more affordable, environmentally friendly, and dependable.

\noindent \textbf{22. Summary\_EcofriendlyConsumerChoices:}
The broader theme of these texts is environmental sustainability and eco-conscious consumer choices. They focus on promoting a laundry detergent product that aligns with sustainability goals by offering a range of environmentally friendly features, such as being cruelty-free, using minimal plastic packaging, and supporting carbon-neutral shipping. The theme encourages individuals to make environmentally responsible choices in their daily lives, particularly in the context of laundry and reducing plastic waste.

\noindent \textbf{23. Summary\_PlasticWasteandEnvironmentalImpact:}
The overall summary of these texts is a collective call to address the urgent issue of plastic waste and pollution in the environment. They highlight the alarming amount of plastic entering our oceans, stress the need for collective action involving both consumers and brands to combat this crisis, introduce innovative solutions like pyrolysis to reduce plastic waste and carbon footprints, and emphasize the importance of better handling and regulation within the petrochemical industry to prevent plastic pollution. The texts underscore the broader theme of environmental responsibility and the imperative to find sustainable alternatives to single-use plastics and improve the entire plastic production and waste management lifecycle.

\noindent \textbf{24. Summary\_PromoteSustainableTransportation:} 
These texts collectively emphasize the importance of electric vehicles (EVs) in combating air pollution and improving air quality. They highlight the role of clean car standards and the accessibility of EVs, particularly in Los Angeles, where local rebate programs make them a cost-effective choice. Shifting from gas-powered vehicles to electric options is promoted for its potential to reduce emissions and decrease exposure to pollution, with a focus on the environmental and health benefits of adopting EVs and the need to take steps to clean the air in Colorado and beyond.

\noindent \textbf{25. Summary\_WaterManagementandSustainability:}
These texts collectively emphasize the discussion of water treatment and billing issues in various cities. The texts touch upon topics such as the cost of building new water treatment plants versus refurbishing existing ones, concerns about water quality and the presence of toxic chemicals in tap water, allegations of potential corruption or kickbacks in decision-making processes, and the financial aspects of water billing, including rates, fees, and expenditures related to water infrastructure. Additionally, some texts emphasize the importance of proper maintenance and infrastructure investment to ensure clean and safe drinking water for residents.

\subsection{COVID-19 Vaccine Campaigns Cluster Summaries}
\label{sec:cl_sum_covid}
Summaries of the final $23$ themes (clusters) from the COVID-19 case study are shown below:

\noindent \textbf{1. Summary\_GovDistrust:} 
The texts express strong criticism towards President Joe Biden and Dr. Anthony Fauci. They accuse President Biden of failing in his duties by mishandling the border and COVID-19 response, causing division within the country, and weakening international alliances. Additionally, Biden's approval ratings are described as suffering due to these issues. Dr. Fauci is accused of lying to Congress and the public, with calls for his imprisonment and accountability for alleged deceptions. The critique extends to the CDC and the Biden administration's policies on vaccinated individuals needing protection from the unvaccinated, questioning the rationale behind these measures.

\noindent \textbf{2. Summary\_GovTrust:} 
Both Donald Trump and President Biden have advocated for COVID-19 vaccination efforts in the U.S., with Trump urging Americans to get vaccinated and Biden actively leading the national vaccination campaign, which includes making all adults vaccine-eligible ahead of schedule. Efforts are also supported by various government levels, ensuring broad access to vaccines. Additionally, Biden's broader administration efforts include rejoining the Paris Agreement, rescinding intended withdrawal from the World Health Organization, and enacting numerous executive orders to tackle the pandemic effectively and enhance public health infrastructure.

\noindent \textbf{3. Summary\_VaccineRollout:} 
COVID-19 vaccines are now FDA approved and available for children aged 5-17, emphasizing protection and safe participation in school and social activities. Health experts, including pediatricians, advocate for vaccination to reduce the risk of severe illness in children. Community outreach includes drive-in events and school system initiatives to facilitate vaccine access. Meanwhile, state and local news platforms promote updates on vaccine rollout, encouraging residents of all ages to get vaccinated, as seen in various U.S. locations including Missouri, West Virginia, and the Bronx.

\noindent \textbf{4. Summary\_VaccineSymptom:}
Following reports of rare blood clots (thrombosis with thrombocytopenia syndrome, TTS) in young adults after receiving the Johnson \& Johnson COVID-19 vaccine, U.S. health agencies recommended a temporary pause to investigate. The European Medicines Agency (EMA) also acknowledged a possible link between the vaccine and unusual blood clots, recommending a warning label. After review, the vaccine was deemed safe, but with caution advised for women aged 18-49 due to a higher risk of TTS. Despite the pause, public health officials have resumed its use, stressing that such adverse events are extremely rare, with only 17 cases reported out of over 8 million doses.

\noindent \textbf{5. Summary\_VaccineEquity:}
Advocacy for equitable COVID-19 vaccine distribution emphasizes the need to prioritize vulnerable and marginalized populations globally, stressing that health care is a human right. Calls for action urge the U.S. and other entities to ensure vaccines are accessible to all, reflecting a widespread belief that no one is safe until everyone is safe. Organizations like Human Rights Watch are holding governments and pharmaceutical companies accountable, while community initiatives continue to support health equity through various programs and events. Overall, the message is clear: overcoming the pandemic requires making vaccines available equally to all individuals worldwide.

\noindent \textbf{6. Summary\_VaccineStatus:}
Local news updates across various counties in the U.S. emphasize the ongoing COVID-19 vaccination efforts. Jefferson County highlights a local doctor promoting vaccination, while Chemung County reports that 30\% of its population is fully vaccinated. Wyandotte County has opened an additional vaccine site to increase accessibility. Alamance County's health department is now taking vaccine appointments via a new hotline, and Mecosta County is providing detailed information on vaccine scheduling. These updates reflect concerted efforts to encourage and facilitate vaccination among the community members.

\noindent \textbf{7. Summary\_EncourageVaccination:}
Across various locations, efforts are being made to facilitate and encourage COVID-19 vaccination. Promotions include no-appointment-needed vaccine clinics, incentives such as free beverages, swimming, and raffle prizes, and convenient settings like the Las Vegas Strip and local community events. Available vaccines include the one-shot Janssen and two-dose options from Pfizer and Moderna, suitable for different age groups. Locations like Miami-Dade County offer easy access at several sites with extended hours and the option to register online or by phone to expedite the process. These initiatives aim to build a healthier community by increasing vaccination rates.

\noindent \textbf{8. Summary\_VaccineMandate:}
There is significant opposition to COVID-19 vaccine mandates and the concept of vaccine passports across various states in the U.S. Critics argue that such requirements infringe on individual freedoms and privacy, posing constitutional concerns. Legal actions and political responses include lawsuits against mandates, state legislation banning vaccine passports, and vocal criticism from public figures who view these measures as an abuse of power. For example, Texas and Florida have passed laws preventing the enforcement of vaccine passports, reflecting a broader national debate on balancing public health needs with personal liberties.

\noindent \textbf{9. Summary\_VaccineReligion:}
Religious perspectives on COVID-19 vaccinations vary, with most major denominations not opposing vaccination, yet some individuals seek religious exemptions. For example, leaders from the South Dakota Catholic Conference acknowledge that Catholics can conscientiously object to vaccines based on religious principles, although they also state that vaccination is not a universal moral obligation. Additionally, Conway Regional Health System addressed an increase in vaccine exemption requests due to concerns about the use of fetal cell lines by providing information on other common medicines developed similarly. Overall, the dialogue reflects a complex interaction between religious beliefs and public health policies.

\noindent \textbf{10. Summary\_VaccineEfficacy:}
The COVID-19 vaccine is broadly acknowledged as effective and essential, particularly highlighted by data showing over 99\% of COVID-related deaths occur among unvaccinated individuals. Healthcare professionals emphasize the vaccine's critical role in reducing infection risks and controlling the pandemic. Pregnant women, advised by their doctors, are also considered safe recipients of the vaccine. This collective understanding among health experts aims to combat hesitancy and encourage widespread vaccination to protect public health.

\noindent \textbf{11. Summary\_VaccineDevelopment:}
The rapid development of COVID-19 vaccines did not compromise safety or efficacy, as they were built on decades of scientific research and established vaccine technology. Innovations like mRNA vaccines were prepped by years of prior study, enabling a swift response to the global pandemic. The FDA's approval underscores that these vaccines meet rigorous standards for safety, effectiveness, and manufacturing quality. Experts like Dr. Sabrina Assoumou note that the urgent global need for a solution to the pandemic catalyzed the mobilization of scientific resources, resulting in these breakthrough vaccines.

\noindent \textbf{12. Summary\_CovidPlan:}
The ongoing COVID-19 pandemic has highlighted societal vulnerabilities, prompting significant legislative and community responses. The American Rescue Plan, passed by Congress, aims to mitigate pandemic effects through increased funding for vaccine distribution and financial support for individuals and families, ensuring resources like childcare providers remain safe and operational. Efforts to expand vaccine availability are evident, with communities like Hudson County receiving significant increases in vaccine allocations. Additionally, initiatives to establish Points of Dispensing (PODs) for vaccines are underway, emphasizing a coordinated approach to safely and efficiently return to normalcy.

\noindent \textbf{13. Summary\_VaccineMisinformation:}
There is no scientific evidence linking COVID-19 vaccines to fertility issues in either men or women. Healthcare professionals and credible sources continue to address and dispel misinformation regarding the impact of vaccination on fertility. The vaccines are deemed safe for those planning pregnancies, and it is highlighted that the vast majority of COVID-19 deaths occur among the unvaccinated. Individuals concerned about vaccine safety are encouraged to consult their doctors and visit reliable health websites for accurate information.

\noindent \textbf{14. Summary\_NaturalImmunity:}
Health experts, including state health officials and CDC heads, are discussing the nuances of immunity in the context of COVID-19, particularly with the emergence of variants like Delta, which may shift herd immunity thresholds. Secretary of Health Kim Malsam-Rysdon suggests that vaccine-induced immunity appears to be more robust than natural immunity. Meanwhile, debates continue regarding the effectiveness of natural immunity versus vaccine-induced immunity, with some Pfizer employees reportedly favoring natural immunity. Public health messaging consistently emphasizes the benefits of vaccination for broader community protection and achieving herd immunity.

\noindent \textbf{15. Summary\_Vote:}
Various messages encourage voting on specific dates, emphasizing the importance of civic participation. Early voting for Robert H in District 6 starts on April 19th, and another vote is scheduled for April 6th, 2021. Additionally, there is a call to vote on November 2nd for several local officials, including a supervisor and town council members. These reminders highlight the opportunity for individuals to influence their community and government by voting and supporting various candidates across different dates and locations.

\noindent \textbf{16. Summary\_VaccineRefusalNews:}
Local news outlets in Lincoln and Cedar Counties in Missouri, as well as Randolph, Logan, and Chicot Counties in Arkansas, are reporting on the number of people who are refusing the COVID-19 vaccine. These reports aim to keep local residents informed about vaccine acceptance rates within their respective communities, highlighting regional responses to the vaccination efforts. The focus on vaccine refusal rates is part of broader coverage to understand and address public health challenges during the pandemic.

\noindent \textbf{17. Summary\_MaskMandate:}
The texts discuss the polarized responses to mask mandates in the U.S. during the COVID-19 pandemic, highlighting varying regional approaches and legal interpretations. Allegany County explicitly rejects mask mandates, emphasizing voluntary vaccination without further restrictions. In contrast, the Illinois State Board of Education, following an executive order, enforces universal indoor masking in schools, citing health safety and legal obligations. A legal victory in New York struck down mask mandates, praised as a significant judicial opinion. Meanwhile, personal opinions express a preference for optional masking, advocating for parental and individual choice over government mandates. These perspectives illustrate the complex interplay between public health directives and individual freedoms during the pandemic.

\noindent \textbf{18. Summary\_VaccineBrewIncentives:}
In Charleston, SC, several breweries are offering incentives to encourage COVID-19 vaccination. Brewlab Charleston, Cooper River Brewing, Tideland Brewing, and Palmetto Brewing Company are all participating in a promotion where individuals can receive a free beverage after getting their vaccine shot at these locations. This initiative aims to increase vaccination rates by combining public health efforts with local business support.

\noindent \textbf{19. Summary\_CommunityServiceByCandidate:}
Candidates across various districts and positions are actively engaging with their communities to address local concerns and improve public services. Victor Ramirez, running for District 2 County Council, highlights his team's achievements like implementing speed humps and hosting vaccination clinics, emphasizing teamwork and community collaboration. Another candidate managed a significant local vaccination effort, booking over 1,500 appointments for residents. In Arlington, an incumbent running for the Select Board discusses challenges and achievements during their term, focusing on education, racial justice, housing affordability, and pandemic response. Each candidate showcases their commitment to addressing specific community needs and improving local governance.

\noindent \textbf{20. Summary\_AdvocatingUnifiedLiberties:}
The texts highlight a strong opposition to government mandates related to COVID-19, including mask-wearing and vaccination requirements. Speakers argue that such mandates infringe on personal freedoms and civil liberties. There is a significant emphasis on the need to protect individual rights and ensure that local and state authorities retain control over public health decisions rather than centralized government. The discourse also touches on moral and spiritual dimensions, suggesting a broader cultural and societal conflict, where respect for diverse viewpoints and human dignity should prevail. Overall, the sentiment is one of resistance to perceived overreach by government authorities and a call to uphold individual liberties.

\noindent \textbf{21. Summary\_CovidEconomy:}
Recent discussions and analyses focus on the economic recovery from the COVID-19 pandemic, highlighting several key elements. In virtual events and forums, experts including Suzanne Clark and Bill Gates have debated the roles of digital solutions and climate change in shaping future economies. Financial commentators and economic historians like Adam Tooze reflect on the lessons from the pandemic, suggesting that the quick adaptation to digital platforms and economic policies have been critical. Reports indicate that while the U.S. shows signs of economic rebound, challenges like inflation, labor shortages, and supply chain issues persist. However, increased vaccination rates and fiscal policies, including stimulus measures, are seen as positive steps towards sustaining economic growth. There's a general consensus that the combination of accommodating monetary policies and strategic fiscal interventions will continue to support the market and economic recovery.

\noindent \textbf{22. Summary\_UrgentPoliticalAdvocacy:}
A Republican senator criticizes Republican governors for promoting vaccine misinformation. A podcast discusses evaluation of news related to Biden’s border policy and woke capitalism. The alleged leader of QAnon announces his campaign run and opponents set up a fund to counteract him. An ongoing opinion poll concerning President Biden and Vice President Harris's administration's approval rate is shared. Forecasts point towards the re-election of New Jersey Governor Phil Murphy, as highlighted by Lawyer-Murphy’s FactWars.com. The summary is derived from various separate texts.

\noindent \textbf{23. Summary\_AgainstSocialistPolicy:}
The texts express opposition to Nancy Pelosi's 'Socialist Drug Takeover Plan', suggesting it would result in government control over the prescription drug market and subsequently fewer breakthrough cures and less innovation. The reader is urged to contact various politicians - Chris Pappas, Josh Gottheimer, Elissa Slotkin, Susan Wild, and Kim Schrier - to voice opposition to Pelosi's plan and specifically to block H.R. 3.
\end{document}